%% file: main.tex
\pdfoutput=1

\documentclass[11pt]{article}

\usepackage[final]{acl}
\usepackage{pgfplots}
\pgfplotsset{compat=1.18} 
\usepackage{subcaption} 
\usepackage{times}
\usepackage{latexsym}
\usepackage[table,xcdraw]{xcolor}
\usepackage{multirow}
\usepackage{adjustbox}
\usepackage{wrapfig}

\usetikzlibrary{positioning,matrix,calc,arrows.meta,decorations.pathreplacing,shapes.geometric}
\definecolor{mygreen}{RGB}{46,139,87}
\definecolor{myred}{RGB}{255,152,150}
\definecolor{myblue}{RGB}{30,144,255}
\definecolor{myyellow}{RGB}{219,219,141}
\definecolor{mybrown}{RGB}{197,157,148}
\usepackage{pgfplotstable}

\usepackage{xspace}

\newcommand{\our}{\textsc{DaID}\xspace}

\usepackage{microtype}
\usepackage{graphicx}
\usepackage{booktabs} 
\usepackage{multirow}
\usepackage[table,xcdraw]{xcolor}

\usepackage{amsmath}
\usepackage{amssymb}
\usepackage{mathtools}
\usepackage{tikz}
\usepackage{pgfplots}
\usepackage{xcolor}

\usepackage{arydshln}
\usepackage{pifont}
\usepackage{colortbl}  
\usepackage{booktabs}
\usepackage[T1]{fontenc}

\usepackage[utf8]{inputenc}

\usepackage{microtype}

\usepackage{inconsolata}

\usepackage{graphicx}

\title{Spotlight and Shadow: Attention-Guided Dual-Anchor Introspective Decoding for MLLM Hallucination Mitigation}

\author{
Yebo Wu$^{1}$\footnotemark[2], Han Jin$^{1}$\footnotemark[2], Zhijiang Guo$^{2}$\footnotemark[1], Li Li$^{1}$\footnotemark[1] \\
$^1$State Key Laboratory of IOTSC, University of Macau \\
$^2$Hong Kong University of Science and Technology (Guangzhou) \\
\texttt{\{yc37926, mc56727, llili\}@um.edu.mo} \\
\texttt{zhijiangguo@hkust-gz.edu.cn} \\
}

\begin{document}
\maketitle
\renewcommand{\thefootnote}{\fnsymbol{footnote}}
\footnotetext[1]{Corresponding authors.}
\footnotetext[2]{Equal contribution.}
\renewcommand{\thefootnote}{\arabic{footnote}}

\begin{abstract}
Multimodal Large Language Models (MLLMs) have demonstrated remarkable reasoning capabilities yet continue to suffer from hallucination, where generated text contradicts visual content. In this paper, we introduce Dual-Anchor Introspective Decoding (\our), a novel contrastive decoding framework that dynamically calibrates each token generation by mining the model's internal perceptual discrepancies. Specifically, \our identifies a Spotlight layer to amplify visual factual signals and a Shadow layer to suppress textual inertia. By leveraging visual attention distributions to guide this dual-anchor selection process, our method ensures precise, token-specific adaptation. Experimental results across multiple benchmarks and MLLMs demonstrate that \our significantly mitigates hallucination while enhancing general reasoning capabilities.

\end{abstract}

\input{section/introduction}
\input{section/Motivation}
\input{section/Method}
\input{section/Experiments}

\input{section/Related_work}

\input{section/conclusion}
\input{section/limitations}

\input{section/Ethics_statement}
\bibliography{custom}

\appendix
\input{section/Appendix}

\end{document}

%% file: section/introduction.tex
\section{Introduction}


Multimodal Large Language Models (MLLMs)~\cite{li2024surveying} have demonstrated exceptional versatility, excelling in tasks ranging from image captioning~\cite{bucciarelli2024personalizing,sarto2025image} to complex reasoning~\cite{yang2025look,xu2026ididtherelarge,xu2026bridging,xu2025videoeraser}. By synergizing the perceptual acuity of vision encoders with the cognitive reasoning capabilities of LLMs, MLLMs achieve unprecedented proficiency in comprehending and generating content grounded in multimodal inputs, fundamentally reshaping the landscape of artificial intelligence~\cite{wu2026tsembed,wu2025elastic}.

Despite these significant strides, the practical deployment of MLLMs remains severely constrained by the persistent issue of hallucination~\cite{zhang2025shallow,zhu2025mitigating,wu2025combating}, where models generate textual descriptions that are unfaithful to the provided visual content. This phenomenon critically undermines the trustworthiness of MLLMs in high-stakes scenarios~\cite{xu2024copyrightmeter,xu2026fraudshield}, such as financial analysis~\cite{gan2024mme} and medical diagnosis~\cite{qiu2024application}. For instance, in medical imaging diagnosis~\cite{alsaad2024multimodal}, a model might hallucinate a non-existent tumor driven by spurious textual correlations, leading to dangerous consequences.


\begin{figure}[t]
  \centering
  \includegraphics[width=0.97\linewidth]{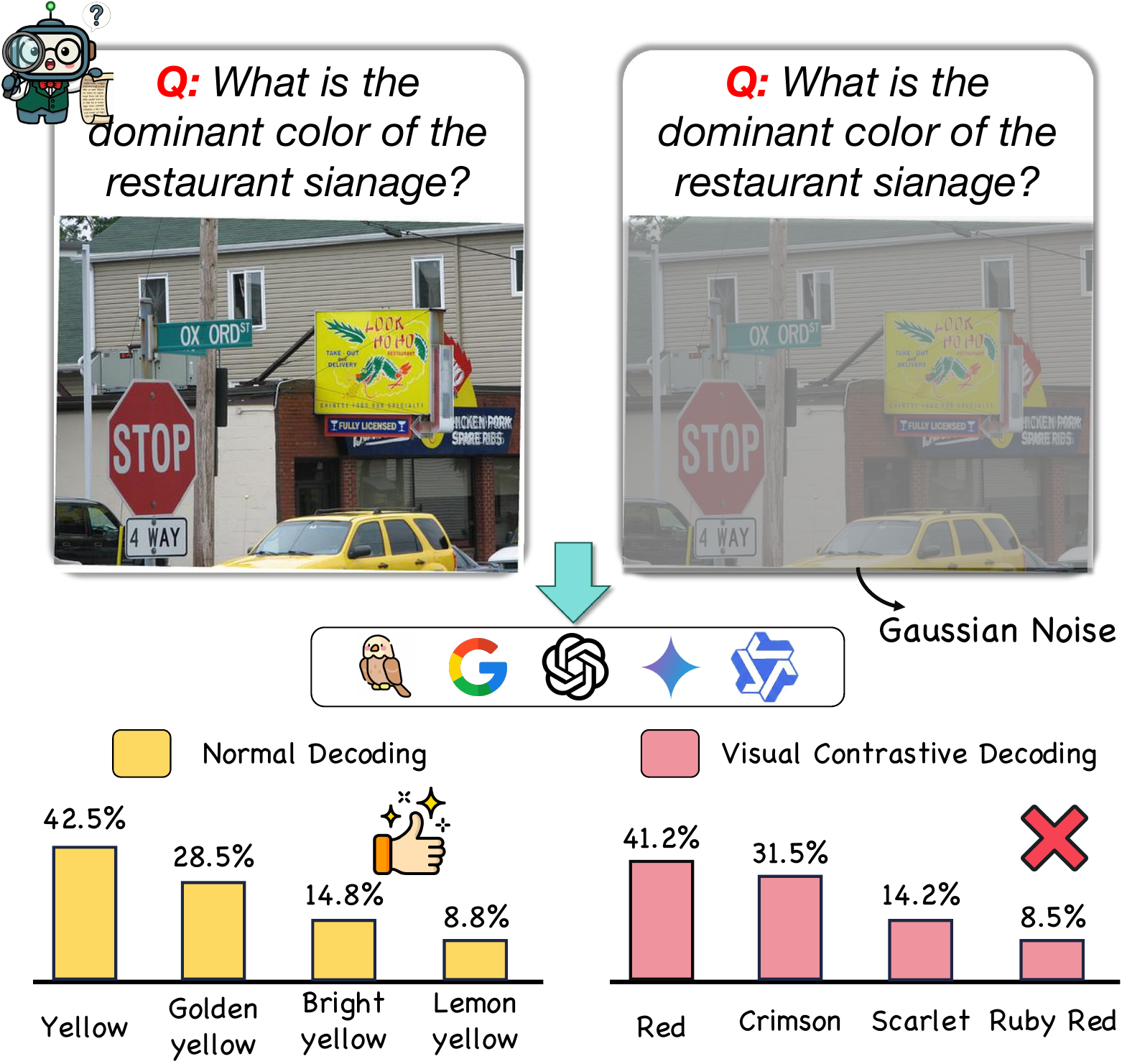}
    \caption{A failure case of Visual Contrastive Decoding (VCD) caused by external perturbations.}
  \label{fig_intro}
  \vspace{-4mm}
\end{figure}

To address this challenge, training-free Contrastive Decoding (CD) strategies have emerged as a promising direction~\cite{wang2024mitigating}. The fundamental premise involves penalizing hallucinated tokens by contrasting the model's logits against a constructed negative distribution~\cite{suo2025octopus}. However, existing CD methods, such as Visual CD (VCD;~\citealt{leng2024mitigating}) and Instruction CD (ICD;~\citealt{wang2024mitigating}), are constrained by two critical limitations. First, these methods necessitate an additional forward pass for the negative sample at each decoding step, imposing substantial computational burden and latency; for instance, VCD increases inference latency by $1.83\times$ compared to standard decoding on NVIDIA V100. Second, existing approaches rely on heuristic external perturbations (e.g., visual masking) to synthesize the negative distribution, which can introduce stochastic noise. As illustrated in Figure~\ref{fig_intro}, such perturbations lead to erroneous semantic shifts: while standard decoding correctly identifies ``yellow'' attributes, VCD inadvertently prioritizes incorrect ``red'' vocabulary due to perturbation-induced noise.


In this paper, we resolve these dilemmas by shifting the paradigm from external intervention to internal introspection. Our approach is grounded in the observation that MLLMs exhibit distinct layer-wise behaviors: shallow layers show severe hallucination tendencies, while intermediate layers possess superior visual perception capabilities. We posit that these internal perceptual discrepancies serve as valuable, intrinsic contrastive sources that can calibrate the generation process. Leveraging this insight, we introduce Dual-Anchor Introspective Decoding (\our), a novel CD method that mines contrastive signals directly from the model's intermediate states. Specifically, \our dynamically identifies two anchors guided by visual attention distributions: a Spotlight layer to amplify factual visual signals and a Shadow layer to suppress linguistic priors. This design enables effective hallucination mitigation within a single forward pass while eliminating the uncertainty noise inherent in traditional methods. Our main contributions are summarized as follows:
\begin{itemize}
    \item We conduct a granular layer-wise diagnosis of MLLM decoding dynamics, revealing that shallow layers exhibit severe hallucination tendencies, whereas intermediate layers possess superior visual perception capabilities.
    
    \item We propose \our, a novel CD method that simultaneously amplifies factual visual signals and suppresses language priors to mitigate hallucinations within a single forward pass.

    \item Our extensive experiments across multiple benchmarks demonstrate that \our significantly outperforms existing methods in hallucination mitigation while enhancing general multimodal reasoning capabilities.

\end{itemize}

%% file: section/Motivation.tex
\section{Motivation and Observation}\label{sec:motivation}

\begin{figure*}[!t]
    \centering
    \begin{subfigure}{0.33\textwidth}
        \centering
        \includegraphics[width=\linewidth]{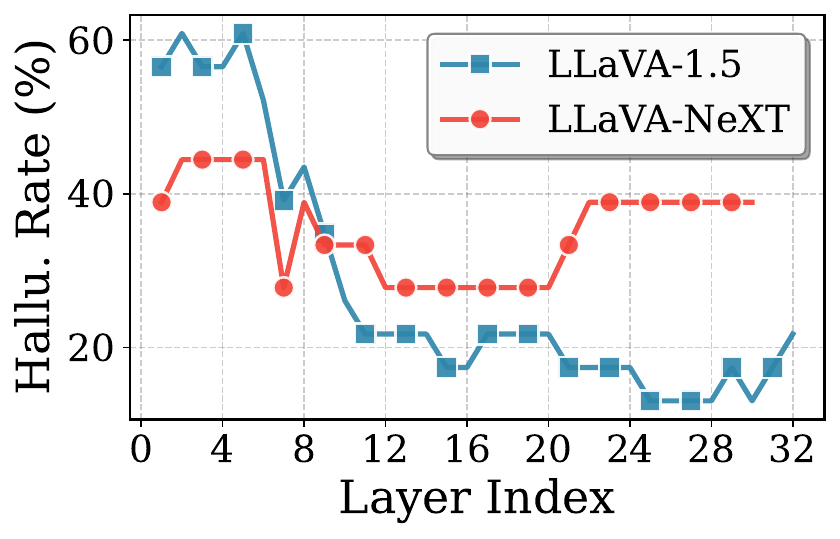}
        \caption{Hallucination Rate.}
    \end{subfigure}\hfill
    \begin{subfigure}{0.33\textwidth}
        \centering
        \includegraphics[width=\linewidth]{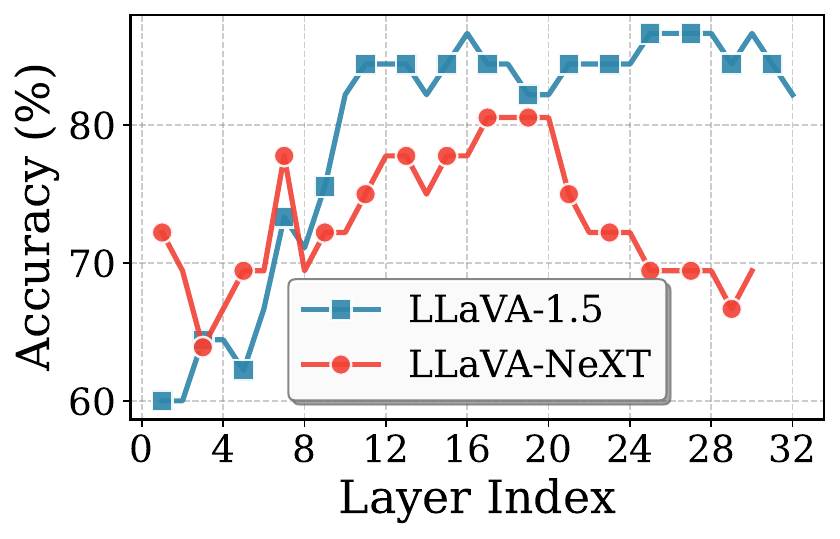}
        \caption{Object Recognition Accuracy.}
    \end{subfigure}\hfill
        \begin{subfigure}{0.33\textwidth}
        \centering
        \includegraphics[width=\linewidth]{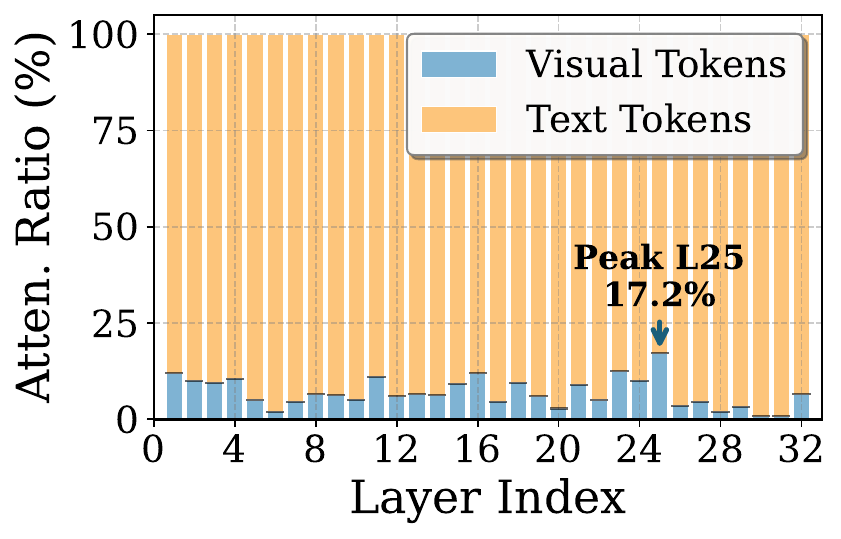}
        \caption{Attention Distribution of LLaVA-1.5.}
    \end{subfigure}
  \caption{Layer-wise diagnosis of MLLM decoding dynamics on the POPE benchmark, revealing the evolution of hallucination rates, visual perception capability, and attention distribution across layers.}
  \label{fig:motivation}
  \vspace{-3mm}
\end{figure*}


In this section, we investigate whether MLLMs possess inherent internal signals necessary to self-correct hallucinations. To answer this, we conduct a layer-wise analysis of the decoding trajectory by probing hidden states and attention distributions across layers. This analysis reveals how visual perception emerges and how hallucination tendencies evolve during generation, thereby identifying optimal internal contrastive anchors for decoding.

\subsection{Evolution of Hallucination Rate}
\label{sec:motivation_shadow}


We first trace the genesis of hallucination by evaluating the factual consistency of layer-wise decoded sequences on the POPE~\cite{li2023evaluating} benchmark. As shown in Figure~\ref{fig:motivation}(a), models exhibit divergent trajectories: LLaVA-1.5~\cite{liu2023visual} shows a generally monotonic decline in hallucination rates, while LLaVA-NeXT~\cite{li2024llavanext-strong} follows a non-monotonic pattern with rates rebounding in deeper layers. Despite these variations, a critical commonality emerges: shallow layers exhibit the most severe hallucination tendencies on both models (peaking at 60.87\% for LLaVA-1.5 and 44.44\% for LLaVA-NeXT), generating fluent but factually detached descriptions.


We attribute this phenomenon to visual agnosia. In initial layers, visual encoder representations are not yet semantically aligned with the LLM's latent space due to the modality gap. While these layers may encode low-level features (e.g., edges, textures), they suffer from a semantic void, lacking the reasoning capability to ground these features into text. This misalignment forces the model to rely on linguistic patterns learned during pre-training, rather than actual visual evidence. Consequently, generation at this stage is driven almost exclusively by unimodal language priors.


\noindent\textbf{\textsc{TakeAway 1:}} Shallow layers, characterized by linguistic noise prevailing over visual reasoning, naturally constitute an ideal ``Shadow'' (negative anchor). This enables us to effectively isolate and subtract the language inertia driving hallucination, leaving clearer visual signals.

\subsection{Evolution of Visual Perception Capability}
\label{sec:motivation_spotlight}

To pinpoint layers where visual comprehension is maximized, we probe object recognition accuracy across the model's depth. As illustrated in Figure~\ref{fig:motivation}(b), recognition accuracy does not follow a simple linear progression. Instead, we identify a distinct seeing-then-forgetting phenomenon, particularly pronounced in LLaVA-NeXT. Specifically, accuracy rises rapidly to a peak at intermediate layers (e.g., layer 25 for LLaVA-1.5 and layer 17 for LLaVA-NeXT), indicating that the model attains optimal visual fidelity within the network's intermediate stages. 
Crucially, this fidelity is not sustained; it degrades significantly in deeper layers, dropping by 4.45\% for LLaVA-1.5 and a striking 11.12\% for LLaVA-NeXT. We attribute this decline to textual inertia, where dominant pre-trained language priors progressively override visual signals to prioritize linguistic fluency and coherence.

\noindent\textbf{\textsc{TakeAway 2:}} 
Intermediate layers, characterized by peak visual fidelity, constitute an ideal ``Spotlight'' (positive anchor). This enables us to amplify authentic perceptual signals, ensuring visually faithful generation.

\subsection{Visual Attention Bridging the Gap}
\label{sec:motivation_attention}
To elucidate the mechanistic underpinnings of the above phenomena and identify these anchors without expensive probes, we employ attention to quantify model's reliance on visual cues. In Figure~\ref{fig:motivation}(c), we visualize attention weights allocated to visual tokens for LLaVA-1.5.
A striking correlation emerges: visual attention reaches its zenith at layer 25, aligning precisely with the layer where object recognition accuracy peaks. This synchronization indicates that high visual attention is a definitive signature of the model's peak perceptual state.


Conversely, in shallower layers (e.g., around layer 6), attention reaches a local minimum, corresponding directly to regions with high hallucination rates.
This validates that visual attention distribution serves as a robust, training-free proxy for the model's cognitive state. Critically, attention weights reveal the intrinsic balance between visual and linguistic processing; this makes them ideal for pinpointing layers with optimal visual grounding, ensuring our dual-anchor selection is driven by the model's actual visual reliance.

\noindent\textbf{\textsc{TakeAway 3:}} Visual attention serves as a faithful, training-free proxy for internal model states. Leveraging this indicator enables token-specific anchoring, precisely identifying the Spotlight and Shadow layers for each decoding step.

%% file: section/Method.tex
\section{Methodology: \our}
\label{sec:method}



\begin{figure*}[t]
  \centering
  \includegraphics[width=1\linewidth]{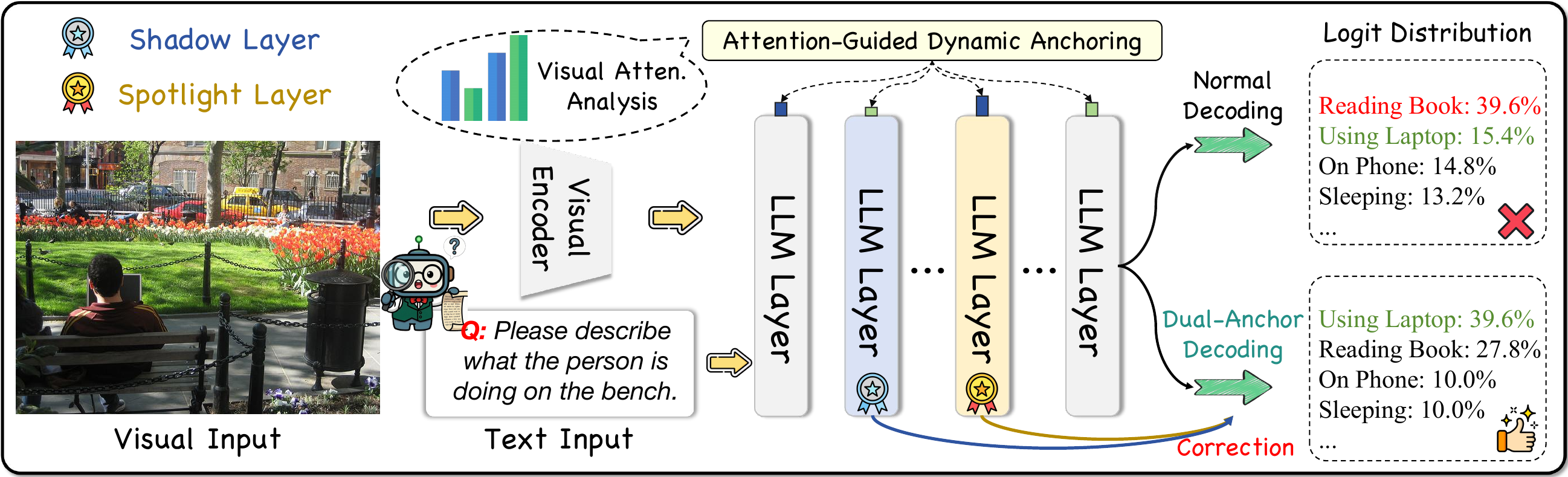}
    \caption{Framework of \our. \our dynamically identifies dual-anchor layers guided by the visual attention and calibrates the final output by leveraging both the Spotlight and Shadow anchors to ensure visual grounding.}
  \label{fig_framework}
  \vspace{-2mm}
\end{figure*}

Drawing on the empirical insights from Section~\ref{sec:motivation}, we introduce \our, a training-free framework designed to mitigate hallucinations. \our dynamically calibrates generation by pinpointing token-specific internal states. As shown in Figure~\ref{fig_framework}, \our employs a Spotlight layer to amplify authentic visual signals and a Shadow layer to suppress linguistic noise.


\subsection{Attention-Guided Dynamic Anchoring}
\label{sec:method_selection}

Our preliminary analysis establishes that the distribution of attention weights serves as a robust proxy for the model's real-time reliance on visual cues. 
To leverage this, we propose the Visual Attention Score (VAS), which dynamically gauges visual reliance at each decoding step.

\paragraph{Visual Attention Score.} Consider the generation of the $t$-th token, where the model processes a context sequence containing $N_v$ visual tokens. Let $A^{l,h}_{t,k} \in [0,1]$ denote the attention weight in layer $l$ and head $h$, representing the affinity between the current query token $t$ and the key token $k$. We define the VAS for layer $l$ at step $t$ as the cumulative attention mass allocated to visual tokens, averaged across all $H$ heads:
\begin{equation}
\label{eq:vas}
\text{VAS}_t(l) = \frac{1}{H} \sum_{h=1}^{H} \sum_{k \in \mathrm{V}} A^{l,h}_{t,k},
\end{equation}
where $\mathrm{V} = \{1, \dots, N_v\}$ represents the set of visual token indices.
Intuitively, a high $\text{VAS}_t(l)$ indicates that layer $l$ actively grounds the current generation in visual features, whereas a low score implies a predominance of linguistic priors.

\paragraph{Spotlight Anchor: Peak Perception.}
To counteract the seeing-then-forgetting phenomenon, where visual signals are attenuated in deeper layers, we identify the positive anchor (Spotlight layer) as the layer exhibiting maximum visual grounding:
\begin{equation}
L_{spot.}^{t} = \operatorname*{argmax}_{l \in \{1, \dots, L\}} \text{VAS}_t(l).
\end{equation}
By anchoring to $L_{spot.}^{t}$, we retrieve authentic visual details that are otherwise diluted in the final layers.

\paragraph{Shadow Anchor: Visual Agnosia.}
Conversely, the negative anchor (Shadow layer) is selected to capture a pre-perception state where visual signals are minimal and linguistic noise dominates. Crucially, we constrain the search space to layers preceding the Spotlight layer to isolate textual inertia before visual features are integrated:
\begin{equation}
L_{shad.}^{t} = \operatorname*{argmin}_{l \in \{1, \dots, L_{spot.}^{t} - 1\}} \text{VAS}_t(l).
\end{equation}
This constraint ($L_{shad.}^{t} < L_{spot.}^{t}$) ensures that the Shadow layer represents pure linguistic noise, providing an ideal state for contrastive suppression.

\subsection{Dual-Anchor Introspective Decoding}
\label{sec:method_decoding}

\our rectifies the final output via a dual-anchor contrastive mechanism. The calibrated logits $\mathcal{L}_{\our}^{t}$ at step $t$ are derived by modulating the final layer's distribution with signals from both anchors:
\begin{align}
\mathcal{L}_{\our}^{t} = &\, [\mathcal{L}_{L_{final}}^{t} + \alpha \cdot \mathcal{L}_{L_{spot.}}^{t}] \cdot (1+\beta) \notag \\
& - \beta \cdot \mathcal{L}_{L_{shad.}}^{t},
\end{align}
where $\alpha$ and $\beta$ modulate the correction intensity, and $\mathcal{L}_{L_{spot.}}^{t}$ and $\mathcal{L}_{L_{shad.}}^{t}$ denote the logits of the Spotlight and Shadow anchors, respectively. This formulation achieves two simultaneous objectives:
\begin{itemize}
    \item \textbf{Visual Perception Reinforcement ($+ \alpha \cdot \mathcal{L}_{L_{spot.}}^{t}$):} Re-injects the peak visual signals to strengthen the model's perceptual acuity.
    \item \textbf{Language Prior Suppression ($- \beta \cdot \mathcal{L}_{L_{shad.}}^{t}$):} Penalizes tokens driven primarily by linguistic correlations to mitigate textual inertia.
\end{itemize}

\paragraph{Adaptive Plausibility Constraint.} Since the Spotlight layer ($L_{spot.}$) typically resides in intermediate stages, directly injecting its logits may introduce syntactically inappropriate tokens despite high visual relevance. Similarly, contrastive subtraction on negligible-probability tokens causes instability. To address these, we restrict dual-anchor contrastive decoding to a candidate set $\mathcal{V}^{'}$, dynamically determined by the final layer's distribution:
\begin{align}
\mathcal{V}^{'}(y_{<t}) &= \{ y_{t} \in \mathcal{V} \mid \, p_{\theta}(y_{t}|\mathcal{L}_{L_{final}},y_{<t}) \notag \\
& \geq \gamma \cdot \max_{w} p_{\theta}(w|\mathcal{L}_{L_{final}},y_{<t}) \},
\end{align}
where $\gamma \in [0, 1]$ confines adjustment to linguistically valid candidates. The dual-anchor adjustment is applied exclusively to candidates within $\mathcal{V}'$:
\begin{equation}
p_{\our}(y_{t}) = \begin{cases}
p_{\our}(y_{t}) & \text{if } y_{t} \in \mathcal{V}^{'}(y_{<t}), \\
0 & \text{otherwise}.
\end{cases}
\end{equation}
This constraint ensures \our sharpens factual details and eliminates inertia while maintaining the generated text's fluency and coherence.

%% file: section/Experiments.tex
\section{Experiments}

\subsection{Experimental Settings}

\paragraph{Baselines.}
We compare \our with the following baselines.
1) DoLa~\cite{chuang2023dola} contrasts early and late layers to isolate factual knowledge; 
2) VCD~\cite{leng2024mitigating} mitigates hallucination by contrasting original and distorted visual inputs; 
3) OPERA~\cite{huang2024opera} employs an attention penalty to prevent token over-trust; 
4) SID~\cite{huo2024self} exploits internal state discrepancies for intrinsic correction; 
and 5) EAZY~\cite{che2025hallucinatory} eliminates hallucinations by zeroing out hallucinatory image tokens.

\paragraph{Evaluation Models.} We validate \our across five representative MLLMs: LLaVA-1.5~\cite{liu2023visual}, LLaVA-NeXT~\cite{li2024llavanext-strong}, Qwen2-VL~\cite{wang2024qwen2}, MiniGPT-4~\cite{zhu2023minigpt}, and InstructBLIP~\cite{dai2023instructblip}. All models are evaluated at the 7B parameter scale.

\paragraph{Implementation Details.} 
We conduct evaluations on NVIDIA RTX 5070 Ti GPUs. For the dual-anchor contrastive decoding, we set $\alpha = 0.8$ to govern the intensity of visual perception reinforcement and $\beta = 0.2$ to modulate the suppression of linguistic priors. Additionally, the confidence threshold for the adaptive plausibility constraint is set to $\gamma = 0.9$ for the POPE benchmark and 0.1 for others, ensuring that the dual-anchor adjustments are confined to a linguistically valid candidate set.

\subsection{Research Questions and Benchmarks}
\label{sec:benchmarks}

We evaluate \our on three visual hallucination and five general vision-language benchmarks to investigate the following research questions:
\begin{itemize}\item 
\textbf{RQ1:} Can \our mitigate hallucinations without sacrificing core reasoning capabilities?
\item \textbf{RQ2:} What is the impact of $\alpha$ (visual reinforcement) and $\beta$ (language suppression)?
\item \textbf{RQ3:} Is \our consistently effective across diverse MLLM architectures?
\item \textbf{RQ4:} What are the individual contributions of the Spotlight and Shadow anchors?
\item \textbf{RQ5:} How does \our qualitatively rectify token distribution during decoding?\end{itemize}

\paragraph{Visual Hallucination Tasks.} These benchmarks assess the factual consistency of model responses.
\begin{itemize}
    \item \textbf{POPE}~\cite{li2023evaluating}: Assesses object hallucination by asking binary questions (e.g., \texttt{``Is there a <object> in the image?''}). We report the Accuracy and F1 score.
    \item \textbf{CHAIR}~\cite{rohrbach2018object}: Evaluates object hallucination in image captioning by comparing generated descriptions against ground-truth annotations. We report the $\text{CHAIR}_I$ (Eq.~\eqref{eq:chair1}) and $\text{CHAIR}_S$ (Eq.~\eqref{eq:chair2}).
    \item \textbf{MME}~\cite{fu2025mme}: Measures perceptual and cognitive abilities. We focus on the object-level (Existence, Count) and attribute-level (Position, Color) subsets, reporting the corresponding perception score.
\end{itemize}
\begin{equation}
    \text{CHAIR}_{\text{I}} = \frac{|\{\text{hallucinated objects}\}|}{\text{all mentioned objects}}.
\label{eq:chair1}
\end{equation}
\begin{equation}
    \text{CHAIR}_{\text{S}}= \frac{|\{\text{captions with hallucinated objects}\}|}{\text{all captions}}.
\label{eq:chair2}
\end{equation}

\paragraph{General Vision-Language Tasks.} These benchmarks evaluate reasoning capabilities of MLLMs.
\begin{itemize}
    \item \textbf{GQA}~\cite{hudson2019gqa}: Focuses on visual reasoning and compositional question answering through image scene graphs.
    \item \textbf{VQA$^{\text{v2}}$}~\cite{goyal2017making}: A standard benchmark requiring open-ended visual question answering grounded in image content.
    \item \textbf{MMB}~\cite{liu2024mmbench}: Evaluates 20 ability dimensions using a circular strategy.
    \item \textbf{Seed$^I$}~\cite{li2023seed}: Assesses generative comprehension and spatial relationship understanding via multiple-choice questions.
    \item \textbf{VizWiz}~\cite{gurari2018vizwiz}: Tests the model's ability to resolve diverse visual queries originating from blind individuals.
\end{itemize}



\input{Table/table_main1}

\input{graph/Figure_other_benchmark}

\subsection{Overall Performance}

To answer \textbf{RQ1}, we evaluate \our's effectiveness in mitigating visual hallucinations while preserving core multimodal reasoning capabilities. 

\paragraph{Visual Hallucination Reduction.} 
Table~\ref{tab:main_results} compares \our with existing methods on LLaVA-1.5 and LLaVA-NeXT across visual hallucination benchmarks. \our consistently outperforms all baselines. On POPE for LLaVA-1.5, \our achieves 85.08\% accuracy and 85.92\% F1-score, improving over baselines by up to 3.7\% and 3.72\%, respectively. On CHAIR, \our achieves the lowest $\text{CHAIR}_S$ (35.9\%) and $\text{CHAIR}_I$ (11.3\%), significantly reducing object hallucinations. \our also delivers optimal performance on MME. Similar gains are observed on LLaVA-NeXT. These results demonstrate \our's effectiveness in mitigating hallucinations through reinforced visual grounding and suppressed linguistic priors.


\paragraph{Reasoning Capability Preservation.} We further evaluate whether \our preserves core multimodal reasoning capabilities across five general vision-language benchmarks. As shown in Figure~\ref{fig:other_bench}, \our not only maintains but consistently enhances performance across 7B and 13B scales of LLaVA-1.5. For instance, on Seed$^{I}$, \our achieves a +2.1\% improvement at the 7B scale. These results indicate that amplifying authentic visual signals while filtering linguistic noise refines the model's representational quality, facilitating robust multimodal understanding beyond error correction.

\subsection{Hyperparameter Analysis}

To answer \textbf{RQ2}, we analyze the impact of $\alpha$  and $\beta$ on \our's performance.

\input{graph/Figure_alpha}
\input{graph/Figure_beta}
\input{graph/Figure_other_MLLM}

\paragraph{Understanding the Visual Perception Reinforcement ($\alpha$).} 
As shown in Figure~\ref{fig:sensitivity_alpha}, performance exhibits an inverted-U pattern across both models, peaking at $\alpha=0.8$. When $\alpha=0.4$, both models achieve relatively low performance (e.g., 83.44\% accuracy for LLaVA-1.5), indicating insufficient reinforcement. As $\alpha$ increases to 0.6 and then to 0.8, performance improves significantly (e.g., LLaVA-1.5 reaches 85.92\% F1 and 85.08\% accuracy), demonstrating that reintroducing visual signals effectively counteracts the seeing-then-forgetting phenomenon. However, performance degrades when $\alpha$ exceeds 0.8: at $\alpha=1.0$, F1 scores drop by 0.71\% and 1.66\%, respectively, while accuracy drops by 0.49\% and 1.70\%, with the decline intensifying at $\alpha=1.2$. This occurs because excessive visual signals from the Spotlight layer overpower the final linguistic modeling, disrupting syntactic coherence.

\paragraph{Understanding the Language Prior Suppression ($\beta$).} 
Figure~\ref{fig:sensitivity_beta} shows that performance peaks at $\beta=0.2$ across both models. When $\beta=0.0$ (no suppression), both models achieve suboptimal performance (e.g., 84.99\% F1 and 84.57\% accuracy for LLaVA-1.5) due to unmitigated linguistic priors. As $\beta$ increases to 0.2, performance improves significantly: LLaVA-1.5 reaches 85.92\% F1 (+0.93\%) and 85.08\% accuracy (+0.51\%), while LLaVA-NeXT achieves 85.76\% F1 (+0.47\%) and 85.32\% accuracy (+0.39\%), demonstrating that a mild penalty neutralizes visual agnosia without harming valid generation. However, performance declines when $\beta$ exceeds 0.2: at $\beta=0.4$, F1 scores drop by 1.51\% and 1.89\%, while accuracy falls by 1.19\% and 1.44\%, respectively. The decline intensifies at $\beta=0.6$, indicating that excessive subtraction over-penalizes the vocabulary space, removing not only hallucinations driven by textual inertia but also valid linguistic connectors.


\subsection{Generalization Analysis}

To answer \textbf{RQ3}, we evaluate \our across diverse MLLMs, including Qwen2-VL, MiniGPT-4, and InstructBLIP. As shown in Figure~\ref{fig:generalization_study}, \our consistently yields significant performance gains across all models and benchmarks.
Specifically, \our markedly suppresses erroneous object generation: $\text{CHAIR}_S$ decreases by 3.6\% for InstructBLIP and 1.9\% for MiniGPT-4 (Figure~\ref{fig:generalization_study}(a)), while $\text{CHAIR}_I$ drops by 5.3\% for InstructBLIP (Figure~\ref{fig:generalization_study}(b)). Beyond error reduction, \our consistently enhances grounding capabilities, achieving F1-score improvements of +1.9\% for InstructBLIP and +0.9\% for Qwen2-VL on POPE (Figure~\ref{fig:generalization_study}(c)). These results demonstrate that our attention-guided dual-anchor mechanism serves as a universal, training-free solution for enhancing MLLM reliability across diverse architectures.

\input{Table/table_ablation}

\begin{figure*}[!t]
    \centering
    \begin{subfigure}{0.505\textwidth}
        \centering
        \includegraphics[width=\linewidth]{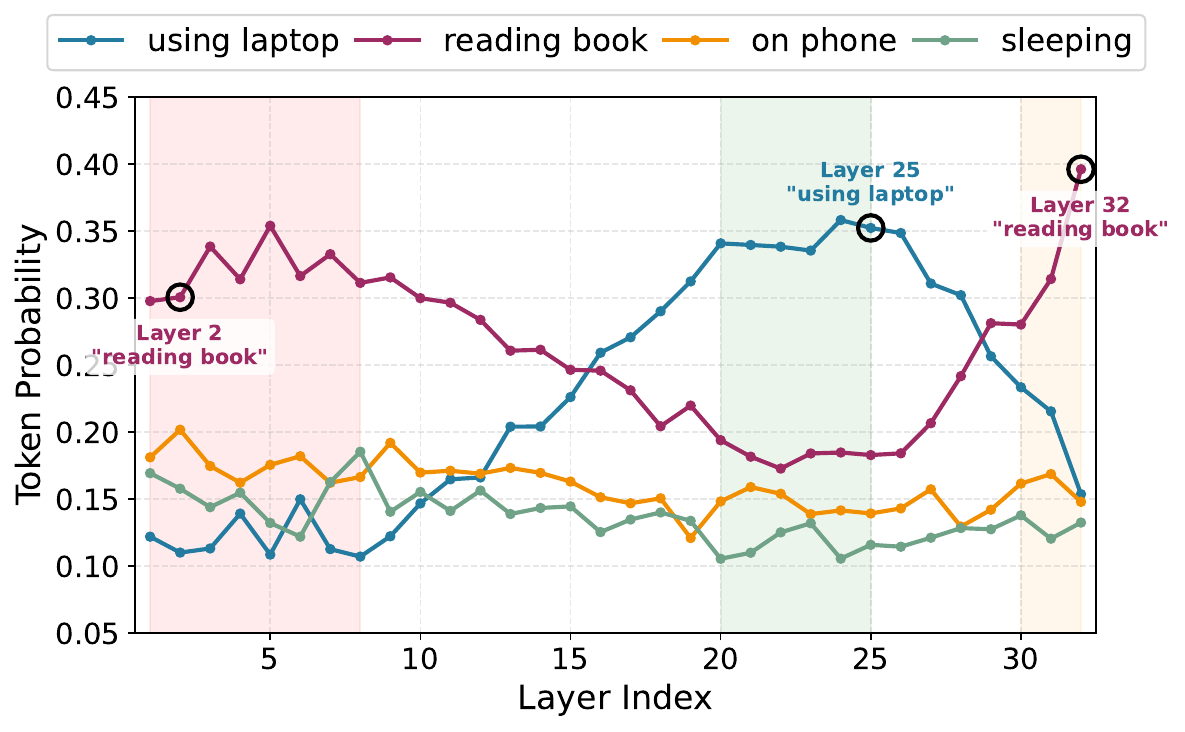}
        \caption{Token probability distribution across layers.}
    \end{subfigure}\hfill
    \begin{subfigure}{0.494\textwidth}
        \centering
        \includegraphics[width=\linewidth]{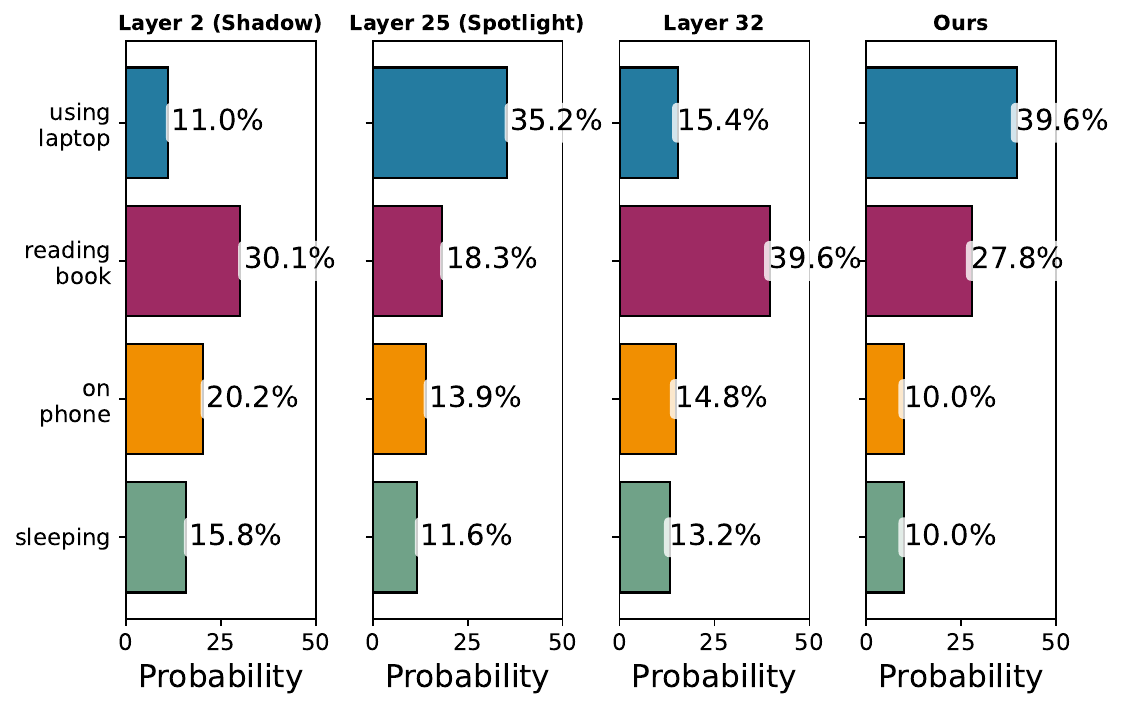}
        \caption{Token probabilities for specific layers.}
    \end{subfigure}
    \caption{Qualitative case study of \our on a representative instance (input image and query detailed in Figure~\ref{fig_framework}).}
  \label{fig_case_study}
  \vspace{-3mm}
\end{figure*}

\subsection{Ablation Study}\label{sec_ablation}

To answer \textbf{RQ4}, we analyze the individual contributions of the Spotlight and Shadow anchors through ablation experiments.
Table~\ref{tab:ablation_results} shows that removing either anchor degrades performance. The Spotlight anchor (Positive Anchor) contributes to visual grounding: excluding it (w/o PA) reduces POPE accuracy from 85.08\% to 84.57\% on LLaVA-1.5, confirming that reinforcing visual perceptual signals is essential to counteract the seeing-then-forgetting phenomenon. The Shadow anchor (Negative Anchor) contributes to hallucination suppression: its absence (w/o NA) causes $\text{CHAIR}_S$ to jump from 35.9\% to 48.9\% on LLaVA-1.5, underscoring its necessity in isolating and penalizing linguistic priors. Similar gains are also observed on LLaVA-NeXT, where both anchors contribute significantly to performance improvements. \our consistently outperforms single-anchor variants across both models, demonstrating that simultaneously amplifying visual signals and suppressing linguistic noise is imperative for effective hallucination mitigation.

\subsection{Qualitative Analysis}

To answer \textbf{RQ5}, we analyze how \our rectifies token distribution during decoding through a representative case where the model erroneously prioritizes ``reading book'' over the visually grounded ``using laptop''. Figure~\ref{fig_case_study}(a) shows that the correct token (``using laptop'') peaks at layer 25 but is suppressed by ``reading book'' at the final layer, illustrating how deeper layers prioritize linguistic fluency over visual evidence. During decoding, \our rectifies this distribution by adaptively designating layer 2 as the Shadow anchor to capture linguistic noise and layer 25 as the Spotlight anchor to amplify visual signals. As shown in Figure~\ref{fig_case_study}(b), \our successfully inverts the erroneous distribution, boosting the correct token's probability from 15.4\% to 39.6\% while suppressing the hallucinated candidate, demonstrating effective token-level rectification during decoding.

%% file: Table/table_main1.tex
\definecolor{steelbluev2}{HTML}{DAE8FC}
\definecolor{steelblue}{HTML}{82B0D2}
\definecolor{color3}{HTML}{FEFAE0}
\newcolumntype{a}{>{\columncolor{green!10}}c}
\newcolumntype{b}{>{\columncolor{blue!10}}c}
\newcolumntype{d}{>{\columncolor{color3}}c}
\begin{table*}[t!]
\centering
\resizebox{\textwidth}{!}{%
\begin{tabular}{lccccccccccc}
\toprule
\multirow{2}{*}{\textbf{Method}} & \multicolumn{2}{c}{\textbf{POPE}} & \multicolumn{2}{c}{\textbf{CHAIR}} & \multicolumn{5}{c}{\textbf{MME}} \\
 \cmidrule(lr){2-3} \cmidrule(lr){4-5} \cmidrule(lr){6-10}
 
 & Acc.$\uparrow$ & F1$\uparrow$ & CHAIR$_S$$\downarrow$ & CHAIR$_I$$\downarrow$ & Existence$\uparrow$ & Count$\uparrow$ & Position$\uparrow$ & Color$\uparrow$ & Total$\uparrow$ \\

\midrule
\midrule
\multicolumn{10}{c}{LLaVA-1.5-7B} \\
\midrule
\midrule
Greedy & 81.38 & 82.20 & 49.6 & 14.4 & 173.45 & 122.72 & 113.39 & 149.92 & 559.48 \\
Beam Search & 84.66 & 84.60 & 46.3 & 12.9 & 175.67 & 124.67 & 114.00 & 151.00 & 565.34 \\

\hdashline

DoLa~\cite{chuang2023dola} & 84.06 & 84.62 & 47.1 & 13.8 & 180.10 & 127.40 & 119.30 & 154.60 & 581.40 \\
VCD~\cite{leng2024mitigating} & 84.66 & 84.52 & 49.2 & 14.8  & \underline{184.66} & \textbf{137.33} & \underline{128.67} & 153.00 & \underline{603.66} \\

OPERA~\cite{huang2024opera} & 84.88 & 85.21 & 45.4 & 12.7 & 180.67 & \underline{133.33} & 111.67 & 123.33 & 549.00 \\

SID~\cite{huo2024self} & 84.82 & 85.50 & 44.2 & 12.2 & 183.90 & 132.20 & 127.80 & \underline{155.90} & 599.80 \\
EAZY~\cite{che2025hallucinatory} & \underline{84.97} & \underline{85.78} & \underline{38.8} & \underline{11.4} & 182.29 & 131.73 & 126.93 & 155.21 & 596.16 \\

\hdashline

\rowcolor{gray!10} \our & \textbf{85.08} & \textbf{85.92} & \textbf{35.9} & \textbf{11.3} & \textbf{186.14} & 130.70 & \textbf{154.46} & \textbf{162.38} & \textbf{633.68}\\

\midrule
\midrule
\multicolumn{10}{c}{LLaVA-NeXT} \\
\midrule
\midrule
Greedy & 83.78 & 82.24 & 32.8 & 9.1 & 179.78 & 127.40 & 121.92 & 151.82 & 580.92 \\

Beam Search & 83.77 & 81.69 & 33.0 & 9.2 & 180.27 & 128.90 & 120.34 & 155.29 & 584.80 \\

\hdashline

DoLa~\cite{chuang2023dola} & 84.53 & 84.77 & 31.3 & 9.0 & 184.45 & 133.07 & 126.72 & 160.38 & 604.62 \\

VCD~\cite{leng2024mitigating} & 83.37 & 83.95 & 32.8 & 8.9 & \underline{187.20} & \textbf{140.16} & \underline{139.91} & 157.41 & \underline{624.68} \\

OPERA~\cite{huang2024opera} & 84.37 & 84.21 & 34.0 & 8.7 & 183.98 & \underline{135.85} & 118.00 & 131.72 & 569.55 \\

SID~\cite{huo2024self} & 84.61 & 85.32 & 32.8 & \underline{8.3} & 185.03 & 134.96 & 137.23 & \underline{161.74} & 618.96 \\

EAZY~\cite{che2025hallucinatory} & \underline{84.91} & \underline{85.40} & \underline{26.8} & \underline{8.3} & 184.55 & 134.18 & 133.52 & 158.89 & 611.14 \\

\hdashline

\rowcolor{gray!10} \our & \textbf{85.32} & \textbf{85.76} & \textbf{24.2} & \textbf{8.2} & \textbf{189.47} & 133.12 & \textbf{156.29} & \textbf{165.52} & \textbf{644.40}\\

\bottomrule
\end{tabular}%
}
\caption{Performance comparison of \our against state-of-the-art methods on POPE, CHAIR, and MME benchmarks. The best results are highlighted in \textbf{bold}, and the second-best results are \underline{underlined}.}
\label{tab:main_results}
\vspace{-3mm}
\end{table*}

%% file: graph/Figure_other_benchmark.tex
\definecolor{red}{RGB}{172,21,28}
\definecolor{blue}{RGB}{39,89,167}
\definecolor{red1}{RGB}{203,104,104}
\definecolor{blue1}{RGB}{104,155,203}

\definecolor{color3}{HTML}{015697}
\definecolor{color2}{HTML}{019496}
\definecolor{color1}{HTML}{FCAEA1}
\definecolor{red}{RGB}{172,21,28}
\definecolor{blue}{RGB}{39,89,167}
\definecolor{red1}{RGB}{203,104,104}
\definecolor{blue1}{RGB}{104,155,203}
\begin{figure*}[t]
\centering
\hspace{-2mm}
\begin{tikzpicture}
    \scriptsize{
  \begin{axis}[
    at={(-22em,-15.5em)},
    anchor=south west,
    ymajorgrids,
    grid style=dashed,
    legend style={at={(0.5,1)}, anchor=south west},
    legend cell align={left},
    ybar,
    enlarge x limits=0.5,
    xtick align=inside,
    height=.23\textwidth,
    width=.23\textwidth,
    bar width=1em,
    xlabel={\scriptsize{(a) GQA.}},
    xlabel style={scale=1.2, yshift=0.2em, xshift=-0.5em},
    ylabel=\footnotesize{\scriptsize Accuracy (\%)},
    ylabel style={scale=1.2, yshift=2.5em},
    symbolic x coords={{1}, {2},},
    xtick=data,
    ymin=60,
    ymax=68,
    ytick={60,64,68},
    nodes near coords align={vertical},
    xticklabels={7B, 13B},
    ylabel style={yshift=-2.7em},
    yticklabel style={/pgf/number format/fixed,/pgf/number format/fixed zerofill,/pgf/number format/precision=0,rotate=0,scale=1.0},
    legend style={yshift=0.8em,xshift=13.5em,font={\tiny},cells={anchor=west},fill opacity=0.8, scale=0.5, legend columns=3, font=\small}
    ]
    \addplot[fill=color1, draw=black, line width=0.8pt, area legend] coordinates {({1},62.0) ({2},63.3)};
    \addlegendentry{\scalebox{1.0}{{Baseline}}}
    \addplot[fill=color2, draw=black, line width=0.8pt, area legend, 
             nodes near coords={$+\pgfmathprintnumber[fixed,precision=1]{\pgfplotspointmeta}$},
             point meta=explicit,
             every node near coord/.style={anchor=south, xshift=-0.3em}] 
             coordinates {({1},63.2)[1.2] ({2},64.6)[1.3]};
    \addlegendentry{\scalebox{1.0}{Baseline + \our}}
  \end{axis}
	
  \begin{axis}[
    at={(-10em,-15.5em)},
    anchor=south west,
    ymajorgrids,
    grid style=dashed,
    legend style={at={(0.02,1)}, anchor=south west},
    legend cell align={left},
    ybar,
    enlarge x limits=0.5,
    xtick align=inside,
    height=.23\textwidth,
    width=.23\textwidth,
    bar width=1em,
    xlabel={\scriptsize{(b) VQA$^{\text{v2}}$.}},
    xlabel style={scale=1.2, yshift=0.2em, xshift=-0.5em},
    ylabel=\footnotesize{\scriptsize Accuracy (\%)},
    ylabel style={scale=1.2, yshift=2.5em},
    symbolic x coords={{1}, {2},},
    xtick=data,
    ymin=75,
    ymax=85,
    ytick={75,80,85},
    nodes near coords align={vertical},
    xticklabels={7B, 13B},
    ylabel style={yshift=-2.7em},
    yticklabel style={/pgf/number format/fixed,/pgf/number format/fixed zerofill,/pgf/number format/precision=0,rotate=0,scale=1.0},
    legend style={yshift=0.2em,xshift=4.2em,font={\tiny},cells={anchor=west},fill opacity=0.8, scale=1.0, legend columns=3}
    ]
    \addplot[fill=color1, draw=black, line width=0.8pt, area legend] coordinates {({1},78.5) ({2},80.0)};
    \addplot[fill=color2, draw=black, line width=0.8pt, area legend,
             nodes near coords={$+\pgfmathprintnumber[fixed,precision=1]{\pgfplotspointmeta}$},
             point meta=explicit,
             every node near coord/.style={anchor=south, xshift=-0.3em}] 
             coordinates {({1},78.9)[0.4] ({2},81.3)[1.3]};

  \end{axis}

  \begin{axis}[
    at={(2em,-15.5em)},
    anchor=south west,
    ymajorgrids,
    grid style=dashed,
    legend style={at={(0.02,1)}, anchor=south west},
    legend cell align={left},
    ybar,
    enlarge x limits=0.5,
    xtick align=inside,
    height=.23\textwidth,
    width=.23\textwidth,
    bar width=1em,
    xlabel={\scriptsize{(c) MMB.}},
    xlabel style={scale=1.2, yshift=0.2em, xshift=-0.5em},
    ylabel=\footnotesize{\scriptsize Accuracy (\%)},
    ylabel style={scale=1.2, yshift=2.5em},
    symbolic x coords={{1}, {2},},
    xtick=data,
    ymin=60,
    ymax=72,
    ytick={60,65,70},
    nodes near coords align={vertical},
    xticklabels={7B, 13B},
    ylabel style={yshift=-2.7em},
    yticklabel style={/pgf/number format/fixed,/pgf/number format/fixed zerofill,/pgf/number format/precision=0,rotate=0,scale=1.0},
    legend style={yshift=0.2em,xshift=4.2em,font={\tiny},cells={anchor=west},fill opacity=0.8, scale=1.0, legend columns=3}
    ]
    \addplot[fill=color1, draw=black, line width=0.8pt, area legend] coordinates {({1},64.3) ({2},67.7)};
    \addplot[fill=color2, draw=black, line width=0.8pt, area legend,
             nodes near coords={$+\pgfmathprintnumber[fixed,precision=1]{\pgfplotspointmeta}$},
             point meta=explicit,
             every node near coord/.style={anchor=south, xshift=-0.3em}] 
             coordinates {({1},65.9)[1.6] ({2},68.9)[1.2]};
  \end{axis}

  \begin{axis}[
    at={(14em,-15.5em)},
    anchor=south west,
    ymajorgrids,
    grid style=dashed,
    legend style={at={(0.02,1)}, anchor=south west},
    legend cell align={left},
    ybar,
    enlarge x limits=0.5,
    xtick align=inside,
    height=.23\textwidth,
    width=.23\textwidth,
    bar width=1em,
    xlabel={\scriptsize{(d) Seed$^{\text{I}}$.}},
    xlabel style={scale=1.2, yshift=0.2em, xshift=-0.5em},
    ylabel=\footnotesize{\scriptsize Accuracy (\%)},
    ylabel style={scale=1.2, yshift=2.5em},
    symbolic x coords={{1}, {2},},
    xtick=data,
    ymin=55,
    ymax=65,
    ytick={55,60,65},
    nodes near coords align={vertical},
    xticklabels={7B, 13B},
    ylabel style={yshift=-2.7em},
    yticklabel style={/pgf/number format/fixed,/pgf/number format/fixed zerofill,/pgf/number format/precision=0,rotate=0,scale=1.0},
    legend style={yshift=0.2em,xshift=4.2em,font={\tiny},cells={anchor=west},fill opacity=0.8, scale=1.0, legend columns=3}
    ]
    \addplot[fill=color1, draw=black, line width=0.8pt, area legend] coordinates {({1},58.6) ({2},61.6)};
    \addplot[fill=color2, draw=black, line width=0.8pt, area legend,
             nodes near coords={$+\pgfmathprintnumber[fixed,precision=1]{\pgfplotspointmeta}$},
             point meta=explicit,
             every node near coord/.style={anchor=south, xshift=-0.3em}] 
             coordinates {({1},60.7)[2.1] ({2},62.4)[0.8]};
  \end{axis}

  \begin{axis}[
    at={(26em,-15.5em)},
    anchor=south west,
    ymajorgrids,
    grid style=dashed,
    legend style={at={(0.02,1)}, anchor=south west},
    legend cell align={left},
    ybar,
    enlarge x limits=0.5,
    xtick align=inside,
    height=.23\textwidth,
    width=.23\textwidth,
    bar width=1em,
    xlabel={\scriptsize{(e) VizWiz.}},
    xlabel style={scale=1.2, yshift=0.2em, xshift=-0.5em},
    ylabel=\footnotesize{\scriptsize Accuracy (\%)},
    ylabel style={scale=1.2, yshift=2.5em},
    symbolic x coords={{1}, {2},},
    xtick=data,
    ymin=48,
    ymax=57,
    ytick={48,51,54,57},
    nodes near coords align={vertical},
    xticklabels={7B, 13B},
    ylabel style={yshift=-2.7em},
    yticklabel style={/pgf/number format/fixed,/pgf/number format/fixed zerofill,/pgf/number format/precision=0,rotate=0,scale=1.0},
    legend style={yshift=0.2em,xshift=4.2em,font={\tiny},cells={anchor=west},fill opacity=0.8, scale=1.0, legend columns=3}
    ]
    \addplot[fill=color1, draw=black, line width=0.8pt, area legend] coordinates {({1},50.0) ({2},53.6)};
    \addplot[fill=color2, draw=black, line width=0.8pt, area legend,
             nodes near coords={$+\pgfmathprintnumber[fixed,precision=1]{\pgfplotspointmeta}$},
             point meta=explicit,
             every node near coord/.style={anchor=south, xshift=-0.3em}] 
             coordinates {({1},51.1)[1.1] ({2},54.3)[0.7]};
  \end{axis}
  
}   
\end{tikzpicture}
\vspace{-3mm}
\caption{Performance evaluation of LLaVA-1.5-7B/13B across five general vision-language benchmarks.}
\label{fig:other_bench}
\vspace{-3mm}
\end{figure*}

%% file: graph/Figure_alpha.tex
\definecolor{red}{RGB}{172,21,28}
\definecolor{blue}{RGB}{39,89,167}
\definecolor{red1}{RGB}{203,104,104}
\definecolor{blue1}{RGB}{104,155,203}
\definecolor{color1}{HTML}{283c63}
\definecolor{color2}{HTML}{00ad7c}

\begin{figure}[!t]
\centering
\hspace{-6mm}
\begin{tikzpicture}
    \scriptsize{
    \begin{axis}
    [
        anchor=north west,
        at={(5em,-5em)},
        ymajorgrids,
        xmajorgrids,
        grid style=dashed,
        width=.25\textwidth,
        height=.20\textwidth,
        yticklabel style={/pgf/number format/precision=0,/pgf/number format/fixed zerofill,scale=1.0},
        xmax=2100,
        xmin=300,
        ymin=83,
        ymax=87,
        xtick={400,800,1200,1600,2000},
        xticklabels={0.4,0.6,0.8,1.0,1.2},
        ytick={83,85,87},
        xlabel={\scriptsize{(a) POPE-F1$\uparrow$.}},
        xlabel style={scale=1.2, yshift=0.2em, xshift=0.1em},
        ylabel=\footnotesize{\scriptsize F1 Score},
        ylabel style={yshift=0.0em, scale=1.2},
        legend style={at={(2.15,1.4)}, anchor=north east, font={\tiny}, cells={anchor=west}, fill opacity=0.8, scale=1.0, legend columns=3}
        ]

        \addplot[red,mark=pentagon*,,mark size=2.5pt,thick,mark options={fill=white,draw=red,line width=1pt}] coordinates {(400,83.63) (800,84.14) (1200,85.92) (1600,85.21) (2000,84.20)};
        \addlegendentry{\scalebox{1.2}{LLaVA-1.5}}

        \addplot[color1,mark=*,mark size=2.5pt,thick,mark options={fill=white,draw=color1,line width=1pt}] coordinates {(400,84.06) (800,85.00) (1200,85.76) (1600,84.10) (2000,84.09)};
        \addlegendentry{\scalebox{1.2}{LLaVA-NeXT}}

    \end{axis}
	
  \begin{axis}
    [
        anchor=north west,
        at={(20em,-5em)},
        ymajorgrids,
        xmajorgrids,
        grid style=dashed,
        width=.25\textwidth,
        height=.20\textwidth,
        yticklabel style={/pgf/number format/precision=0,/pgf/number format/fixed zerofill,scale=1.0},
        xmax=2100,
        xmin=300,
        ymin=83,
        ymax=86,
        xtick={400,800,1200,1600,2000},
        xticklabels={0.4,0.6,0.8,1.0,1.2},
        ytick={83,84,85,86},
        xlabel={\scriptsize{(b) POPE-Acc$\uparrow$.}},
        xlabel style={scale=1.2, yshift=0.2em, xshift=0.1em},
        ylabel=\footnotesize{\scriptsize Accuracy (\%)},
        ylabel style={yshift=0.em, scale=1.2},
        legend style={at={(0.5,1.2)}, anchor=north east, font={\tiny}, cells={anchor=west}, fill opacity=0.8, scale=1.0, legend columns=3}
        ]

        \addplot[red,mark=pentagon*,,mark size=2.5pt,thick,mark options={fill=white,draw=red,line width=1pt}] coordinates {(400,83.44) (800,83.98) (1200,85.08) (1600,84.59) (2000,83.72)};

        \addplot[color1,mark=*,mark size=2.5pt,thick,mark options={fill=white,draw=color1,line width=1pt}] coordinates {(400,84.24) (800,84.93) (1200,85.32) (1600,84.25) (2000,83.78)};
    \end{axis}}   
    \end{tikzpicture}
    \vspace{-2mm}
    \caption{Impact of visual perception reinforcement intensity $\alpha$ on LLaVA-1.5 and LLaVA-NeXT.}
    \label{fig:sensitivity_alpha}
    \vspace{-4mm}
\end{figure}

%% file: graph/Figure_beta.tex
\definecolor{red}{RGB}{172,21,28}
\definecolor{blue}{RGB}{39,89,167}
\definecolor{red1}{RGB}{203,104,104}
\definecolor{blue1}{RGB}{104,155,203}
\definecolor{color1}{HTML}{283c63}
\definecolor{color2}{HTML}{00ad7c}

\begin{figure}[!t]
\centering
\hspace{-6mm}
\begin{tikzpicture}
    \scriptsize{
    \begin{axis}
    [
        anchor=north west,
        at={(5em,-5em)},
        ymajorgrids,
        xmajorgrids,
        grid style=dashed,
        width=.25\textwidth,
        height=.20\textwidth,
        yticklabel style={/pgf/number format/precision=0,/pgf/number format/fixed zerofill,scale=1.0},
        xmax=1700,
        xmin=300,
        ymin=81,
        ymax=87,
        xtick={400,800,1200,1600},
        xticklabels={0.0,0.2,0.4,0.6},
        ytick={82,84,86},
        xlabel={\scriptsize{(a) POPE-F1$\uparrow$.}},
        xlabel style={scale=1.2, yshift=0.2em, xshift=0.1em},
        ylabel=\footnotesize{\scriptsize F1 Score},
        ylabel style={yshift=0.0em, scale=1.2},
        legend style={at={(2.15,1.4)}, anchor=north east, font={\tiny}, cells={anchor=west}, fill opacity=0.8, scale=1.0, legend columns=3}
        ]

        \addplot[red,mark=pentagon*,,mark size=2.5pt,thick,mark options={fill=white,draw=red,line width=1pt}] coordinates {(400,84.99) (800,85.92) (1200,84.03) (1600,82.17)};
        \addlegendentry{\scalebox{1.2}{LLaVA-1.5}}

        \addplot[color1,mark=*,mark size=2.5pt,thick,mark options={fill=white,draw=color1,line width=1pt}] coordinates {(400,85.29) (800,85.76) (1200,84.25) (1600,82.37)};
        \addlegendentry{\scalebox{1.2}{LLaVA-NeXT}}

    \end{axis}
	
  \begin{axis}
    [
        anchor=north west,
        at={(20em,-5em)},
        ymajorgrids,
        xmajorgrids,
        grid style=dashed,
        width=.25\textwidth,
        height=.20\textwidth,
        yticklabel style={/pgf/number format/precision=0,/pgf/number format/fixed zerofill,scale=1.0},
        xmax=1700,
        xmin=300,
        ymin=81,
        ymax=86,
        xtick={400,800,1200,1600},
        xticklabels={0.0,0.2,0.4,0.6},
        ytick={81,83,85},
        xlabel={\scriptsize{(b) POPE-Acc$\uparrow$.}},
        xlabel style={scale=1.2, yshift=0.2em, xshift=0.1em},
        ylabel=\footnotesize{\scriptsize Accuracy (\%)},
        ylabel style={yshift=0.em, scale=1.2},
        legend style={at={(0.5,1.2)}, anchor=north east, font={\tiny}, cells={anchor=west}, fill opacity=0.8, scale=1.0, legend columns=3}
        ]

        \addplot[red,mark=pentagon*,,mark size=2.5pt,thick,mark options={fill=white,draw=red,line width=1pt}] coordinates {(400,84.57) (800,85.08) (1200,83.89) (1600,82.01)};

        \addplot[color1,mark=*,mark size=2.5pt,thick,mark options={fill=white,draw=color1,line width=1pt}] coordinates {(400,84.93) (800,85.32) (1200,83.88) (1600,82.12)};
    \end{axis}}   
    \end{tikzpicture}
    \vspace{-2mm}
    \caption{Impact of language prior suppression intensity $\beta$ on LLaVA-1.5 and LLaVA-NeXT.}
    \label{fig:sensitivity_beta}
    \vspace{-4mm}
\end{figure}

%% file: graph/Figure_other_MLLM.tex
\definecolor{red}{RGB}{172,21,28}
\definecolor{blue}{RGB}{39,89,167}
\definecolor{red1}{RGB}{203,104,104}
\definecolor{blue1}{RGB}{104,155,203}

\definecolor{color3}{HTML}{015697}

\definecolor{color1}{RGB}{200,200,200}  
\definecolor{color2}{RGB}{30,99,184}    
\definecolor{color1}{RGB}{176,196,222}  
\definecolor{color2}{RGB}{13,148,136}   
\definecolor{color1}{RGB}{253,180,98}   
\definecolor{color2}{RGB}{46,134,171}   

\definecolor{red}{RGB}{172,21,28}
\definecolor{blue}{RGB}{39,89,167}
\definecolor{red1}{RGB}{203,104,104}
\definecolor{blue1}{RGB}{104,155,203}
\begin{figure*}[t]
\centering
\hspace{-2mm}
\begin{tikzpicture}
    \scriptsize{
  \begin{axis}[
    at={(-22em,-15.5em)},
    anchor=south west,
    ymajorgrids,
    grid style=dashed,
    legend style={at={(0.5,1)}, anchor=south west},
    legend cell align={left},
    ybar,
    enlarge x limits=0.5,
    xtick align=inside,
    height=.23\textwidth,
    width=.3\textwidth,
    bar width=1em,
    xlabel={\scriptsize{(a) CHAIR$_{S}$$\downarrow$.}},
    xlabel style={scale=1.2, yshift=0.2em, xshift=-0.5em},
    ylabel=\footnotesize{\scriptsize CHAIR$_{S}$},
    ylabel style={scale=1.2, yshift=2.5em},
    symbolic x coords={{1}, {2}, {3}},
    xtick=data,
    ymin=20,
    ymax=70,
    ytick={20,30,40,50,60,70},
    nodes near coords align={vertical},
    xticklabels={Qwen2-VL, MiniGPT-4, InstructBLIP},
    xticklabel style={rotate=30, anchor=north east},
    ylabel style={yshift=-2.7em},
    yticklabel style={/pgf/number format/fixed,/pgf/number format/fixed zerofill,/pgf/number format/precision=0,rotate=0,scale=1.0},
    legend style={yshift=0.8em,xshift=8.5em,font={\tiny},cells={anchor=west},fill opacity=0.8, scale=0.5, legend columns=3, font=\small}
    ]
    \addplot[fill=color1, draw=black, line width=0.8pt, area legend,
             nodes near coords={$-\pgfmathprintnumber[fixed,precision=2]{\pgfplotspointmeta}$},
             point meta=explicit,
             every node near coord/.style={anchor=south, xshift=-0.3em}] 
             coordinates {({1},25.0)[1.4] ({2},31.8)[1.9] ({3},58.8)[3.6]};
    \addlegendentry{\scalebox{1.0}{{Baseline}}}
    \addplot[fill=color2, draw=black, line width=0.8pt, area legend] 
             coordinates {({1},23.6) ({2},29.9) ({3},55.2)};
    \addlegendentry{\scalebox{1.0}{Baseline + \our}}
  \end{axis}
	
  \begin{axis}[
    at={(-2em,-15.5em)},
    anchor=south west,
    ymajorgrids,
    grid style=dashed,
    legend style={at={(0.02,1)}, anchor=south west},
    legend cell align={left},
    ybar,
    enlarge x limits=0.5,
    xtick align=inside,
    height=.23\textwidth,
    width=.3\textwidth,
    bar width=1em,
    xlabel={\scriptsize{(b) CHAIR$_{I}$$\downarrow$.}},
    xlabel style={scale=1.2, yshift=0.2em, xshift=-0.5em},
    ylabel=\footnotesize{\scriptsize CHAIR$_{I}$},
    ylabel style={scale=1.2, yshift=2.5em},
    symbolic x coords={{1}, {2}, {3}},
    xtick=data,
    ymin=5,
    ymax=30,
    ytick={5,10,15,20,25,30},
    nodes near coords align={vertical},
    xticklabels={Qwen2-VL, MiniGPT-4, InstructBLIP},
    xticklabel style={rotate=30, anchor=north east},
    ylabel style={yshift=-2.7em},
    yticklabel style={/pgf/number format/fixed,/pgf/number format/fixed zerofill,/pgf/number format/precision=0,rotate=0,scale=1.0},
    legend style={yshift=0.2em,xshift=4.2em,font={\tiny},cells={anchor=west},fill opacity=0.8, scale=1.0, legend columns=3}
    ]
    \addplot[fill=color1, draw=black, line width=0.8pt, area legend,
             nodes near coords={$-\pgfmathprintnumber[fixed,precision=2]{\pgfplotspointmeta}$},
             point meta=explicit,
             every node near coord/.style={anchor=south, xshift=-0.3em}] 
             coordinates {({1},7.3)[1.0] ({2},9.9)[1.8] ({3},23.7)[5.3]};
    \addplot[fill=color2, draw=black, line width=0.8pt, area legend] 
             coordinates {({1},6.3) ({2},8.1) ({3},18.4)};

  \end{axis}

  \begin{axis}[
    at={(18em,-15.5em)},
    anchor=south west,
    ymajorgrids,
    grid style=dashed,
    legend style={at={(0.02,1)}, anchor=south west},
    legend cell align={left},
    ybar,
    enlarge x limits=0.5,
    xtick align=inside,
    height=.23\textwidth,
    width=.3\textwidth,
    bar width=1em,
    xlabel={\scriptsize{(c) POPE-F1$\uparrow$.}},
    xlabel style={scale=1.2, yshift=0.2em, xshift=-0.5em},
    ylabel=\footnotesize{\scriptsize F1},
    ylabel style={scale=1.2, yshift=2.5em},
    symbolic x coords={{1}, {2}, {3}},
    xtick=data,
    ymin=60,
    ymax=95,
    ytick={60,70,80,90},
    nodes near coords align={vertical},
    xticklabels={Qwen2-VL, MiniGPT-4, InstructBLIP},
    xticklabel style={rotate=30, anchor=north east},
    ylabel style={yshift=-2.7em},
    yticklabel style={/pgf/number format/fixed,/pgf/number format/fixed zerofill,/pgf/number format/precision=0,rotate=0,scale=1.0},
    legend style={yshift=0.2em,xshift=4.2em,font={\tiny},cells={anchor=west},fill opacity=0.8, scale=1.0, legend columns=3}
    ]
    \addplot[fill=color1, draw=black, line width=0.8pt, area legend] coordinates {({1},86.6) ({2},70.3)  ({3},84.4)};
    \addplot[fill=color2, draw=black, line width=0.8pt, area legend,
             nodes near coords={$+\pgfmathprintnumber[fixed,precision=1]{\pgfplotspointmeta}$},
             point meta=explicit,
             every node near coord/.style={anchor=south, xshift=-0.3em}] 
             coordinates {({1},87.5)[0.9] ({2},71.1)[0.8] ({3},86.3)[1.9]};
  \end{axis}

}   
\end{tikzpicture}
\vspace{-1mm}
\caption{Generalization analysis of \our across diverse MLLM architectures on CHAIR and POPE benchmarks.}
\label{fig:generalization_study}
\vspace{-3mm}
\end{figure*}

%% file: Table/table_ablation.tex
\begin{table}[t]
\centering
\resizebox{\columnwidth}{!}{
\renewcommand{\arraystretch}{1.} 
\begin{tabular}{lccccccccccc}
\toprule
\multirow{2}{*}{Method} & \multicolumn{2}{c}{\textbf{POPE}} & \multicolumn{2}{c}{\textbf{CHAIR}} \\
\cmidrule(lr){2-3} \cmidrule(lr){4-5} \cmidrule(lr){6-6} 
 & Acc.$\uparrow$ & F1$\uparrow$ & CHAIR$_S$$\downarrow$ & CHAIR$_I$$\downarrow$ \\
\midrule
\multicolumn{12}{l}{\textbf{\textit{LLaVA-1.5}}} \\
\rowcolor[gray]{0.95} \hspace{1em} + \textsc{\our} & \textbf{85.08} & \textbf{85.92} & \textbf{35.9} & \textbf{11.3} \\
\hspace{1em} w/o PA & 84.57 & 84.99 & 38.1 & 12.0 \\
\hspace{1em} w/o NA & 82.79 & 82.85 & 48.9 & 14.3\\
\midrule
\multicolumn{12}{l}{\textbf{\textit{LLaVA-NeXT}}} \\
\rowcolor[gray]{0.95} \hspace{1em} + \textsc{\our} & \textbf{85.32} & \textbf{85.76} & \textbf{24.2} & \textbf{8.2}\\
\hspace{1em} w/o PA & 84.93 & 85.29 & 26.7 & 8.4 \\
\hspace{1em} w/o NA & 83.11 & 83.52 & 33.6 & 9.1 \\
\bottomrule
\end{tabular}
}
\caption{Ablation study of \our. PA and NA denote the positive and negative anchors, respectively.}
\label{tab:ablation_results}
\vspace{-3mm}
\end{table}

%% file: section/Related_work.tex
\section{Related Work}

\paragraph{Hallucination in MLLMs.} 
Hallucination in LLMs is typically defined as the generation of factually incorrect yet syntactically plausible text~\cite{ji2023towards,xu2026agents}, and this phenomenon is further compounded in MLLMs where the objective shifts from maintaining internal consistency to ensuring cross-modal fidelity~\cite{bai2024hallucination, zhao2025guiding}. Despite architectural advancements like InstructBLIP's Q-Former~\cite{dai2023instructblip}, MLLMs often exhibit a perceptual-cognitive paradox, prioritizing linguistic plausibility over visual evidence~\cite{huo2024self, chuang2023dola}. Recent studies characterize this as degenerative alignment, where visual signals diminish through deeper layers and eventually succumb to linguistic priors~\cite{wang2024mllm}, ultimately triggering hallucinations that favor probabilistic token sequences over visual grounding.


\paragraph{Contrastive Decoding for Hallucination Mitigation.}
To mitigate hallucinations without re-training, Contrastive Decoding (CD)~\cite{li2023contrastive} has emerged as a predominant paradigm. Existing methods fall into two categories: external methods like VCD~\cite{leng2024mitigating} and ICD~\cite{wang2024mitigating} require manual construction of contrastive samples and additional forward passes, introducing extraneous noise and substantial computational overhead, while internal methods like SID~\cite{huo2024self} and DeCo~\cite{wang2024mllm} rely on heuristics or fixed layer selection, failing to adapt to token-specific patterns. In contrast, \our leverages visual attention as a dynamic proxy to adaptively identify dual anchors, enabling token-aware hallucination mitigation without manual intervention or fixed heuristics.

%% file: section/conclusion.tex
\section{Conclusion}

This paper introduces \our, a novel contrastive decoding framework that mitigates MLLM hallucinations by adaptively identifying dual anchors—a Spotlight layer to amplify visual signals and a Shadow layer to suppress linguistic priors—ensuring visually grounded token generation. Experimental results demonstrate that \our significantly reduces hallucination rates while enhancing general multimodal reasoning capabilities.

%% file: section/limitations.tex
\section*{Limitations}
Despite the significant improvements in hallucination mitigation and reasoning efficiency, our work has several limitations that warrant further exploration.
First, computational hardware constraints limited our evaluation to models at the 7B parameter scale. While \our consistently demonstrates superior performance, its scalability and internal layer dynamics in ultra-large-scale MLLMs (e.g., those exceeding 70B parameters) remain to be fully investigated. However, given that the seeing-then-forgetting phenomenon is a fundamental characteristic of Transformer-based cross-modal alignment, we expect our dual-anchor mechanism to generalize effectively to larger architectures. Second, our current investigation primarily focuses on image-text multimodal tasks. While \our effectively mitigates hallucinations by mining internal perceptual discrepancies in image-based MLLMs, its applicability to other modalities, such as video or audio, has not yet been explored. Since temporal dynamics in video understanding introduce more complex attention shifts over time, extending the Dual-Anchor Introspective Decoding framework to handle sequential multimodal data would be a promising direction for future research.

%% file: section/Ethics_statement.tex
\section*{Ethical Considerations}

In this work, we present \our to enhance the factual consistency of MLLMs. All experiments were conducted using publicly available, standard benchmarks (e.g., POPE, CHAIR, MME) that do not contain sensitive personal information or violate privacy. We strictly follow the licenses of all datasets and models used in this work, ensuring compliance with their terms of use and intended purposes. Our method is designed to improve the factual consistency of vision-language models in their intended application domains, and we do not advocate for any use that could cause harm, violate privacy, or be inconsistent with the intended use of the underlying models and datasets.

By reducing hallucinations where generated text contradicts visual content, our method promotes the development of more reliable and honest AI systems for high-stakes scenarios. As a training-free decoding strategy, \our does not introduce new training data that could exacerbate existing societal or demographic biases inherent in pre-trained models. Furthermore, \our operates within a single forward pass, significantly reducing computational overhead and carbon footprint compared to traditional contrastive decoding methods. We acknowledge potential risks: while \our reduces visual-textual inconsistencies, it may not eliminate all forms of hallucinations or errors, and users should exercise appropriate caution in critical applications.

\section*{Acknowledgements}

This work is supported in part by the Science and Technology Development Fund of Macau (0107/2024/RIA2, 0061/2025/RIB2), Joint Science and Technology Research Project with Hong Kong and Macau in Key Areas of Nansha District's Science and Technology Plan (EF2024-00180-IOTSC) and the Multi-Year Research Grant of University of Macau (MYRG-GRG2023-00211-IOTSC-UMDF, MYRG-GRG2024-00180-IOTSC).

%% file: section/Appendix.tex
\section{Discussion and Analysis}

\begin{figure*}[!t]
    \centering
    \begin{subfigure}{0.4\textwidth}
        \centering
        \includegraphics[width=\linewidth]{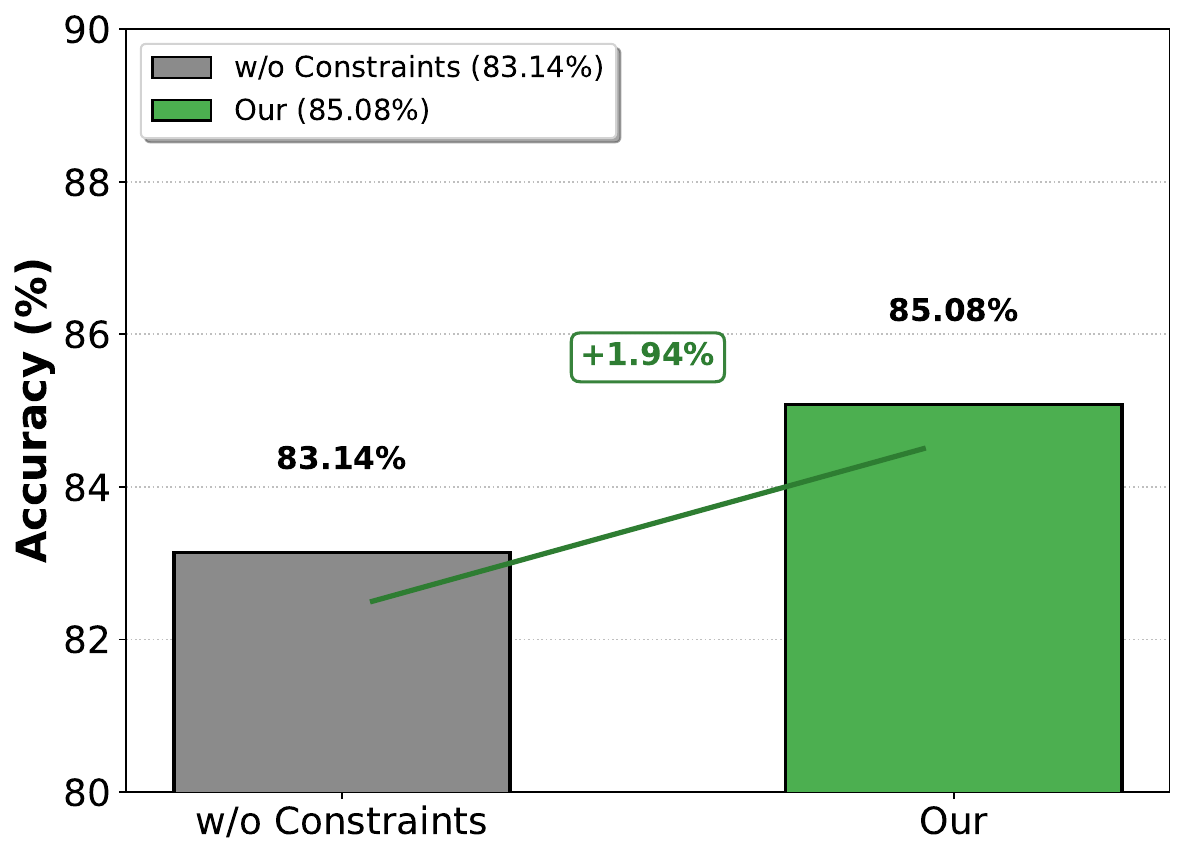}
        \caption{POPE-Acc.}
    \end{subfigure}
    \begin{subfigure}{0.4\textwidth}
        \centering
        \includegraphics[width=\linewidth]{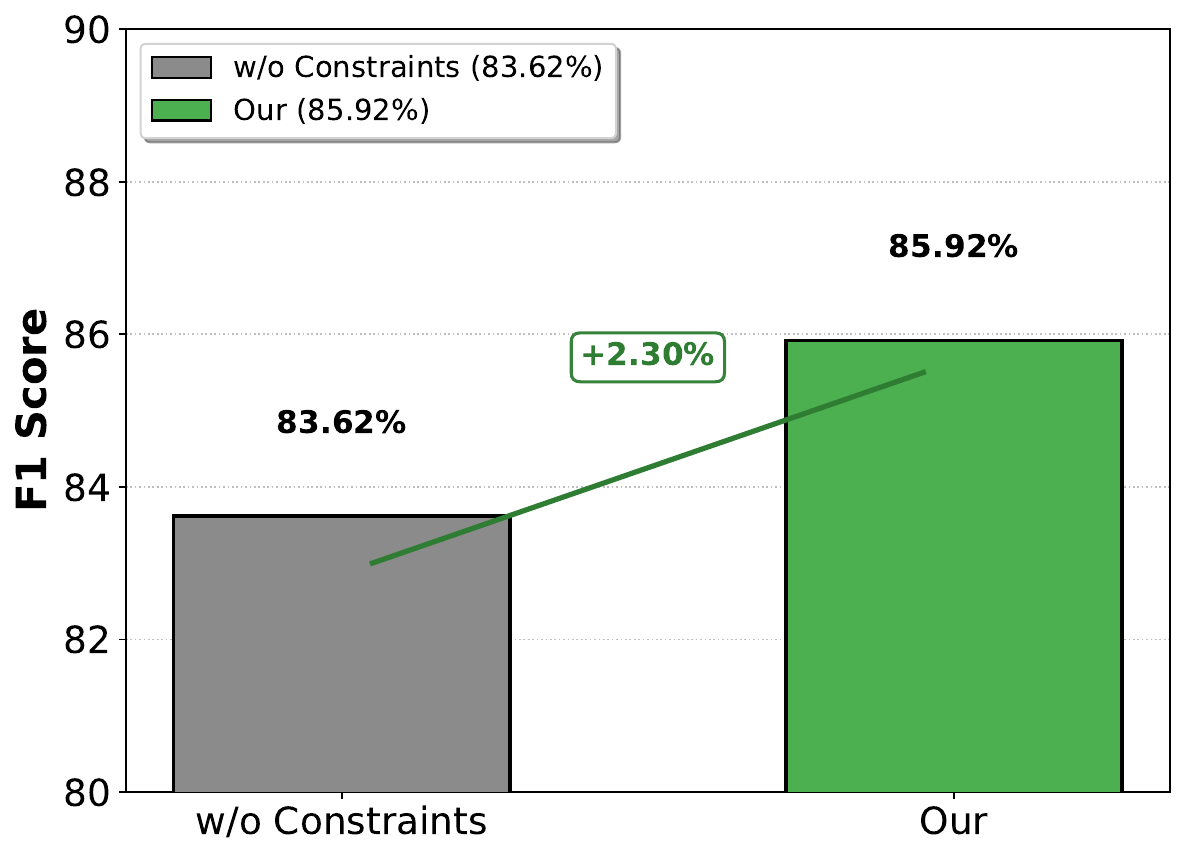}
        \caption{POPE-F1.}
    \end{subfigure}
    \caption{Analysis of the topological constraint ($L_{shad.} < L_{spot.}$) on LLaVA-1.5-7B. We compare the performance between the constrained and unconstrained settings.}
\label{fig_constraints_analysis}
\vspace{-3mm}
\end{figure*}

\subsection{Averaging Attention for Anchor Selection}



Our method computes the Visual Attention Score (VAS) by averaging attention patterns across all heads within each layer. While individual heads may exhibit diverse behaviors, we adopt this layer-level strategy based on three key justifications:

\begin{itemize}
    \item \textbf{Alignment with Decoding Granularity.} As the decoding process operates on layer representations—which aggregate outputs from all heads—selecting anchors at the layer level ensures structural consistency with the model's natural information flow~\cite{wu2026developmental,wu2025memory}. The head-averaged VAS effectively captures the collective visual grounding, mirroring how multi-head information is integrated during inference.

    \item \textbf{Robustness via Consensus.} Averaging across heads distills a consensus visual attention signal while mitigating spurious patterns from individual heads. This robustness is critical for our dynamic, token-specific selection mechanism, ensuring stability across diverse generation contexts~\cite{zhang2026coupled,zhang2025subspace}.

    \item \textbf{Empirical Effectiveness.} 
    Despite head-level variations, our VAS successfully identifies layers with peak visual perception (Section~\ref{sec:motivation}) and delivers significant performance gains across all benchmarks, demonstrating that the consensus signal effectively captures the essential visual grounding information needed for anchor selection.

\end{itemize}

\begin{figure*}[!t]
    \centering
    \begin{subfigure}{0.499\textwidth}
        \centering
        \includegraphics[width=\linewidth]{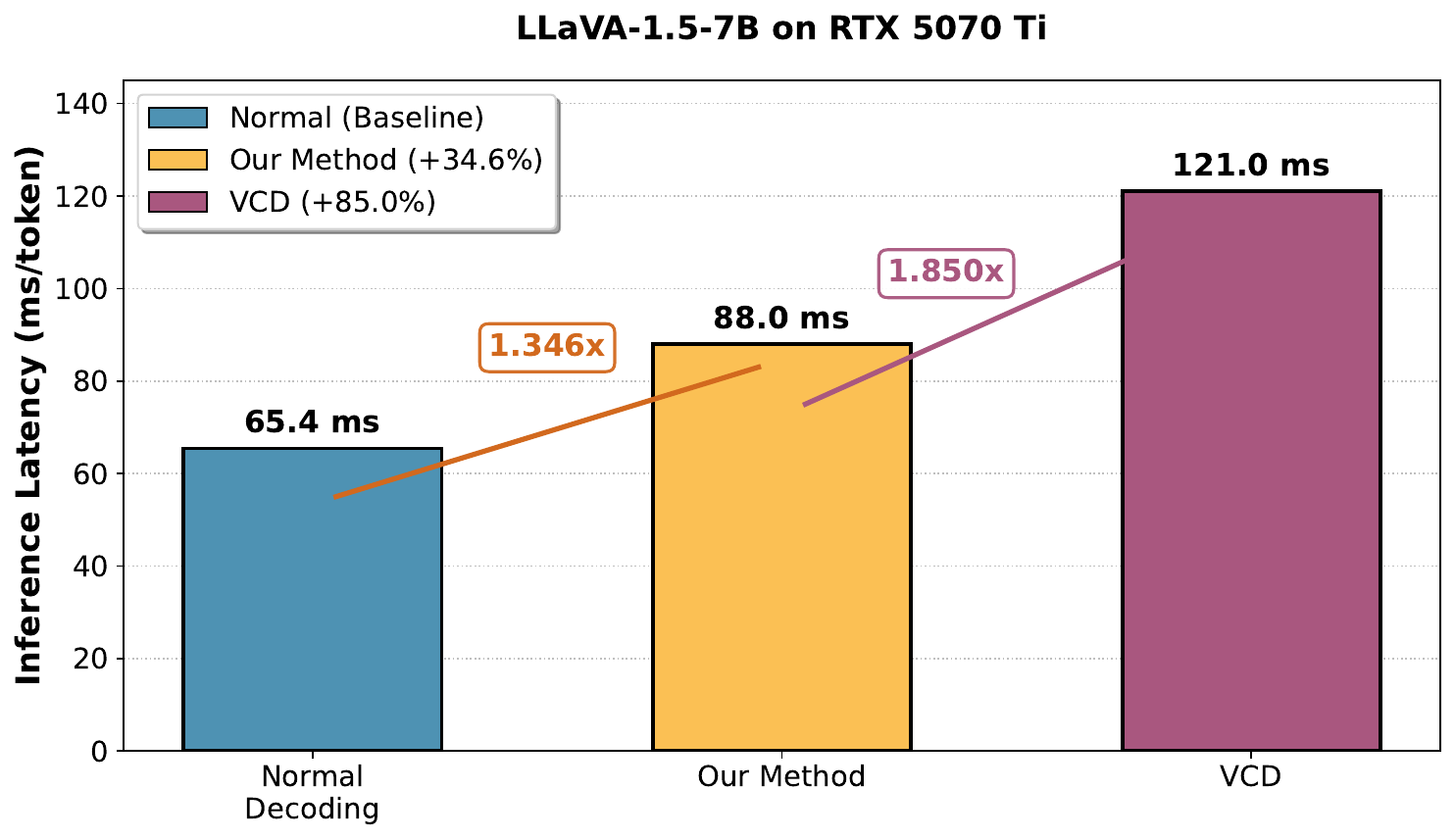}
        \caption{Inference Latency on LLaVA-1.5-7B.}
    \end{subfigure}\hfill
    \begin{subfigure}{0.499\textwidth}
        \centering
        \includegraphics[width=\linewidth]{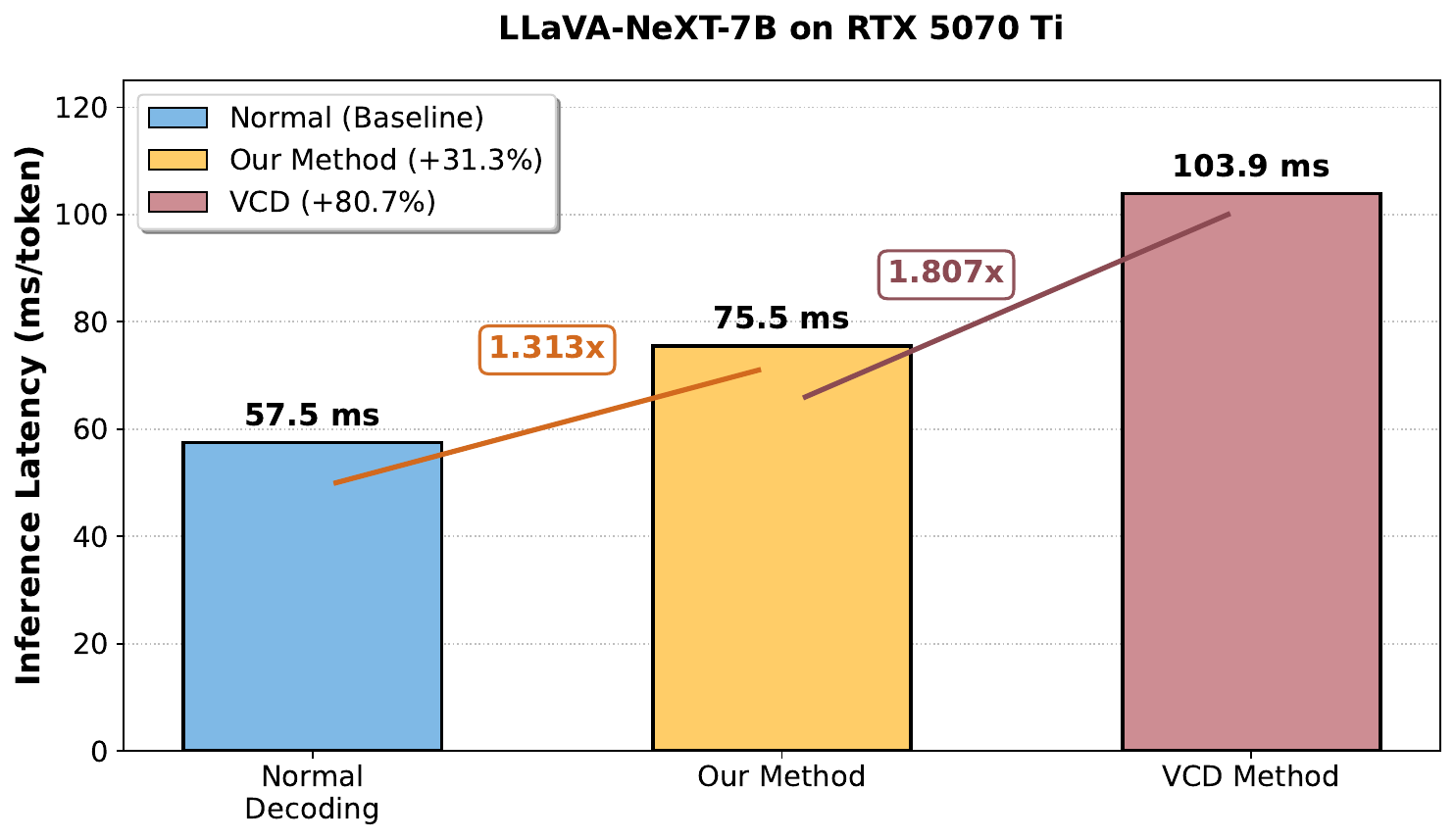}
        \caption{Inference Latency on LLaVA-NeXT-7B.}
    \end{subfigure}
    \caption{Inference latency comparison (ms/token) on NVIDIA RTX 5070 Ti. We evaluate the decoding speed of normal decoding (Baseline), our method, and VCD on LLaVA-1.5-7B and LLaVA-NeXT-7B.}
  \label{fig_time_compare}
  \vspace{-3mm}
\end{figure*}

\subsection{Constraints on Shadow Layer Selection}
\label{sec:constraint_analysis}

We enforce a topological constraint requiring the Shadow layer to precede the Spotlight layer ($L_{shad.} < L_{spot.}$). This design is grounded in both theoretical rationale and empirical evidence.


\begin{itemize}
    \item \textbf{Conceptual Rationale.} This constraint guarantees that the Shadow layer captures a pre-perceptual state~\cite{wu2026endtoenddynamicchainoptimization,wu2025breaking}, where visual signals are minimal and linguistic priors dominate. This ordering aligns with the model's internal progression from visual agnosia (in shallow layers) to peak perception (in intermediate layers). By enforcing this precedence, we isolate pure linguistic noise rather than the semantic drift found in deeper layers, which is critical for maximizing the efficacy of contrastive decoding.

    \item \textbf{Empirical Validation.} Figure~\ref{fig_constraints_analysis} demonstrates the necessity of this topology: relaxing the restriction (``w/o Constraints'') noticeably degrades performance, whereas applying our constraint boosts POPE accuracy from 83.14\% to 85.08\% (+1.94\%) and F1 from 83.62\% to 85.92\% (+2.30\%). These results confirm that anchoring the negative reference to shallow, visually agnostic layers is essential for effective hallucination mitigation.

\end{itemize}

\begin{figure*}[!t]
    \centering
    \begin{subfigure}{0.499\textwidth}
        \centering
        \includegraphics[width=\linewidth]{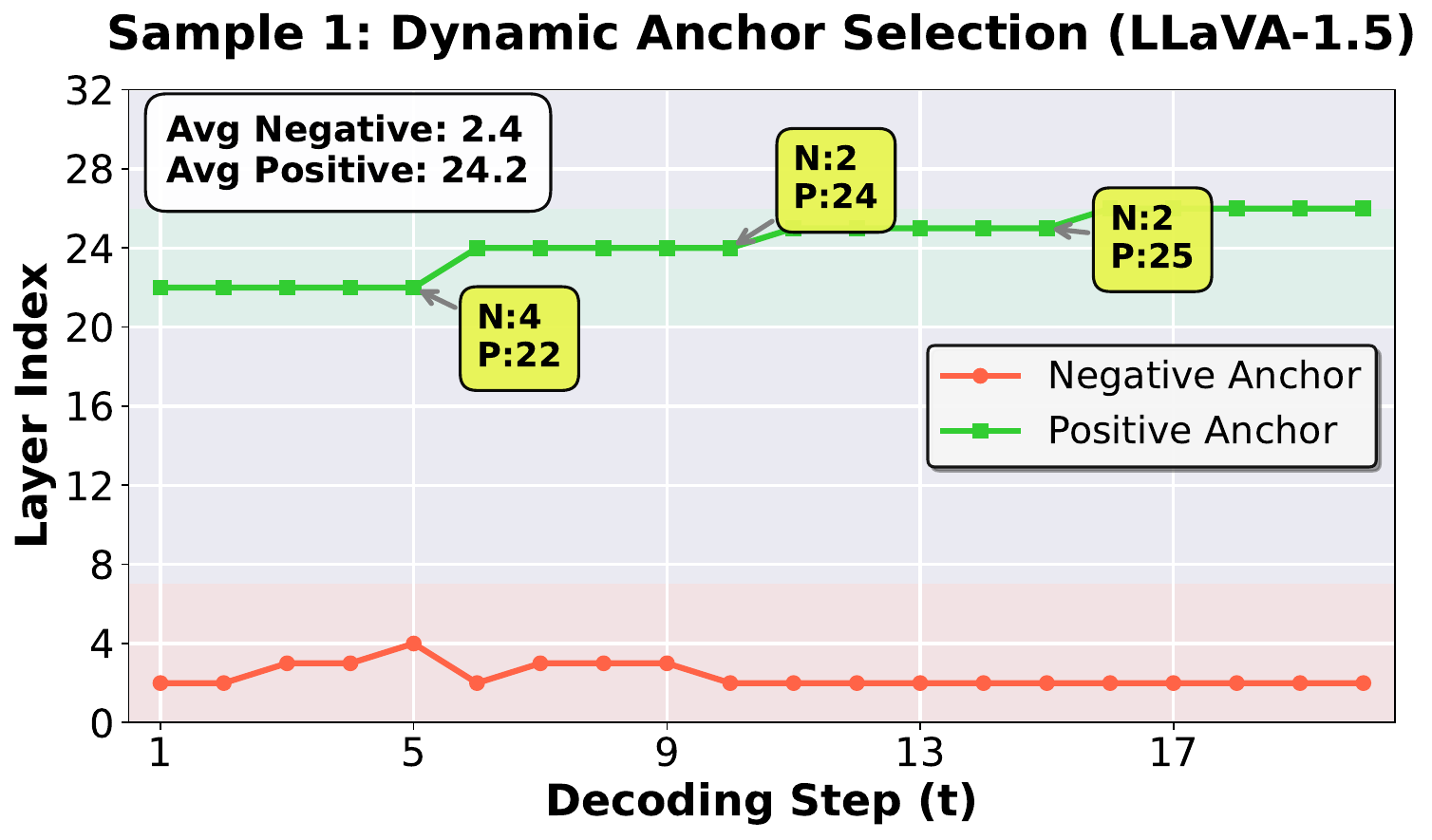}
        \caption{Sample 1 (Stable Progression).}
    \end{subfigure}\hfill
    \begin{subfigure}{0.499\textwidth}
        \centering
        \includegraphics[width=\linewidth]{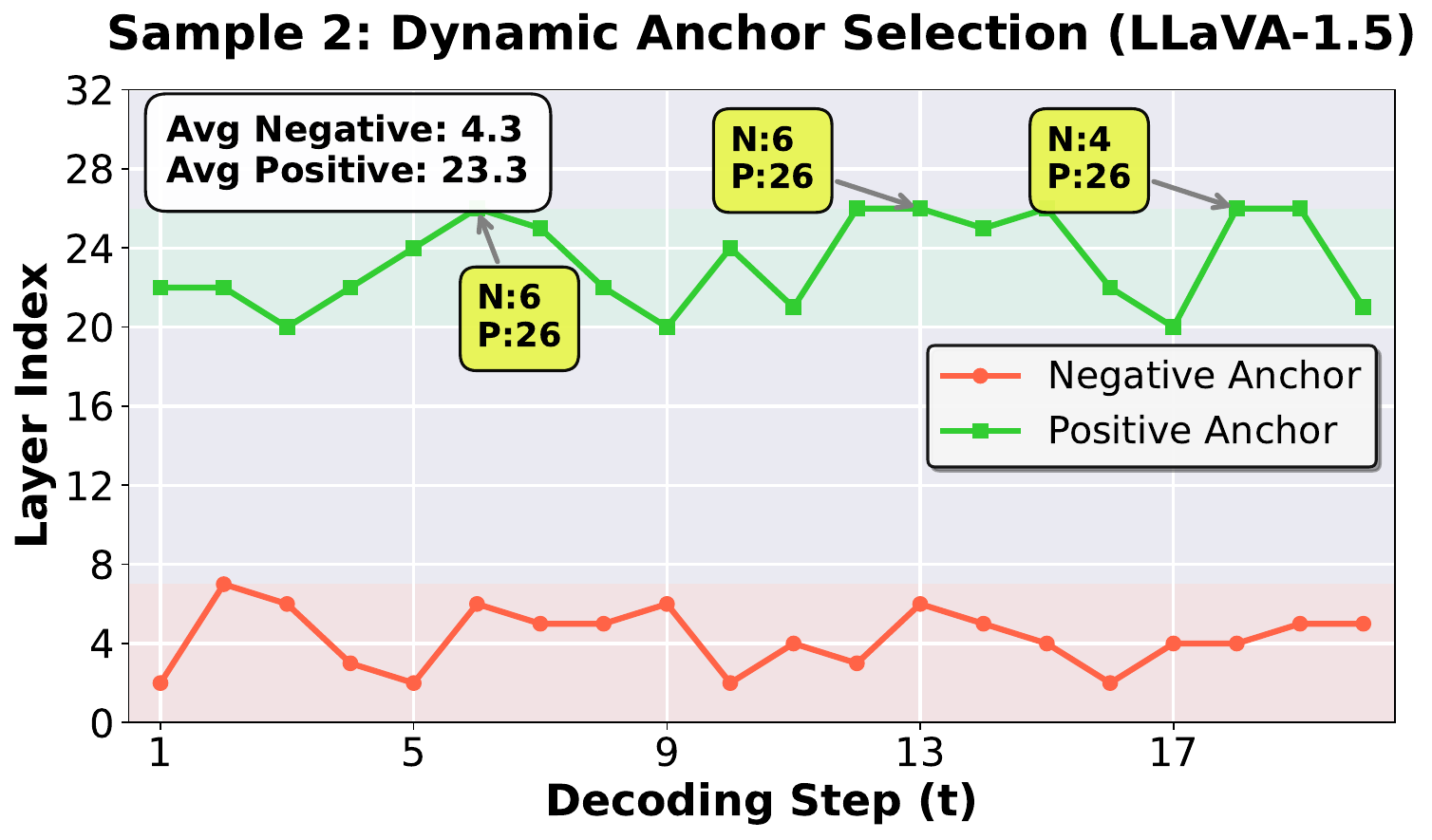}
        \caption{Sample 2 (High Variability).}
    \end{subfigure}
    \caption{Visualization of token-specific dynamic anchor selection on LLaVA-1.5. We track the layer indices selected for the Spotlight Anchor and Shadow Anchor across 20 decoding steps.}
  \label{fig_layer_selection}
\end{figure*}

\subsection{Analysis of Computational Efficiency}

Our approach introduces minimal computational overhead compared to standard decoding while maintaining significantly higher efficiency than external perturbation-based methods (e.g., VCD).

\begin{itemize}
    \item \textbf{Theoretical Analysis.} Standard decoding computes attention weights and hidden states for all layers in a single forward pass. \our incurs only two minor sources of overhead: 1) averaging attention weights for VAS (negligible as it reuses intermediate computations), and 2) projecting the hidden states of the selected Shadow and Spotlight layers to the vocabulary space. Crucially, this design circumvents the computational bottleneck of methods like VCD, which necessitate a second full forward pass on perturbed inputs. Consequently, \our maintains a single-pass inference regime ($\approx 1\times$ cost), whereas VCD doubles the computational cost ($\approx 2\times$).

    \item \textbf{Practical Efficiency.} We evaluate inference latency on an NVIDIA RTX 5070 Ti (Figure~\ref{fig_time_compare}). \our incurs moderate overhead (approx. $1.31\text{--}1.35\times$) compared to the baseline, primarily due to multi-layer logit projections. However, this is significantly more efficient than VCD, which increases latency by over $1.8\times$. On LLaVA-NeXT, \our achieves \textbf{75.5 ms/token}, 37.6\% faster than VCD (103.9 ms/token), demonstrating that mining internal states is effective and efficient.
\end{itemize}

\subsection{Rationale for $\gamma$ Selection}

The adaptive plausibility constraint utilizes distinct $\gamma$ values to align with the vocabulary distributions of different benchmarks. For discriminative benchmarks like POPE, the probability mass is heavily concentrated on a few tokens (e.g., Yes/No). A high threshold ($\gamma=0.9$) ensures that the constraint remains tight, focusing only on the most confident predictions. In contrast, open-ended tasks (e.g., CHAIR) exhibit dispersed probability distributions across a wide vocabulary. A lower threshold ($\gamma=0.1$) allows the method to operate on a wider plausible search space, balancing diversity with the suppression of implausible tokens.


\subsection{Token-Specific Anchor Selection}

The variability in anchor choices across tokens in our dynamic selection mechanism is beneficial and well-controlled through several mechanisms.

\begin{itemize}
    \item \textbf{Context-Dependent Adaptation.} 
    Layer selection is highly responsive to context, reflecting varying perceptual requirements. As shown in Figure~\ref{fig_layer_selection}(a), tokens requiring consistent visual grounding exhibit natural smoothness. Conversely, Figure~\ref{fig_layer_selection}(b) illustrates that when processing complex visual semantics, the model demonstrates adaptive agility, shifting anchors (e.g., oscillating between layers 20 and 26) to calibrate specifically for individual tokens. This fluctuation reflects \our's fine-grained responsiveness.


    \item \textbf{Operational Zone Stability.} Despite index-level fluctuations, the functional roles of anchors remain consistent. As visualized by the shaded regions in Figure~\ref{fig_layer_selection}, the Shadow Anchor is confined to the visual agnosia zone (shallow layers, indices 0--7), while the Spotlight Anchor operates in the peak perception zone (intermediate layers, indices 20--26). This ensures that the contrastive signal, suppressing linguistic noise and amplifying visual features, remains stable throughout generation, regardless of specific layer indices.

    \item \textbf{Linguistic Coherence Guarantee.} The adaptive plausibility constraint (Section~\ref{sec:method}) safeguards coherence by restricting decoding adjustments to the linguistically valid candidate set $\mathcal{V}'$. Consequently, even if internal layer selections vary dynamically, the final output tokens remain syntactically and semantically consistent with the preceding context.

\end{itemize}

\subsection{Metric for Anchor Selection}

We choose Visual Attention Score (VAS) as our  metric for layer selection, motivated by several key considerations over alternative internal signals.

\begin{itemize}
    \item \textbf{Direct Interpretability.} VAS serves as a direct, training-free proxy for visual reliance, explicitly measuring the attention weight assigned to image tokens. This avoids the semantic entanglement often present in hidden state representations.
    

    \item \textbf{Computational Efficiency.} VAS is computationally free, utilizing attention maps generated during standard forward passes. It avoids the latency overhead associated with computing complex vector similarities or gradients.


    \item \textbf{Empirical Validation.} As detailed in Section~\ref{sec:motivation}, VAS strongly correlates with visual perception accuracy. The alignment between peak VAS and optimal object recognition layers confirms its reliability.

\end{itemize}

\subsection{Impact of Anchor Cardinality}
Our method employs two anchors (Spotlight and Shadow), balancing effectiveness and efficiency through complementary mechanisms.

\begin{itemize}
    \item \textbf{Synergistic Roles.} Ablation studies (Section~\ref{sec_ablation}) confirm that both anchors are indispensable: the Spotlight anchor amplifies visual evidence (+0.51\% accuracy), while the Shadow anchor attenuates linguistic priors (-13\% CHAIR$_S$). Removing either component degrades performance, validating that coupling positive visual reinforcement with negative linguistic suppression is essential for effective mitigation.

    \item \textbf{Mechanistic Rationale.}
    The dual-anchor design aligns with contrastive learning principles: amplifying positive signals (visual grounding) while suppressing negative signals (linguistic noise). This configuration avoids the insufficiency of single-anchor methods while circumventing the diminishing returns associated with multi-anchor complexity.


\end{itemize}

\subsection{Robustness Analysis}

While motivated by the seeing-then-forgetting phenomenon, \our is not strictly bound to its intensity. The method operates on the relative differential of visual attention ($\Delta \text{VAS}$) rather than absolute layer indices. Even in instances where the visual attenuation is subtle (a flat forgetting curve), our dynamic selection maximizes the available contrast by locating the optimal local peaks and valleys. 
This property, combined with the universality of attention mechanisms, ensures robustness across varied architectures and diverse input complexities without requiring heuristic tuning.

\section{The Use of Large Language Models}
In the preparation of this manuscript, we utilized LLMs exclusively for linguistic refinement and editing to improve the clarity and readability of the text. We affirm that the conceptualization, methodology, experimental design, and analysis presented in this work are entirely original and were not generated by AI tools. All modifications suggested by the LLMs were critically reviewed and verified by the authors.